\pdfoutput=1

\documentclass[11pt]{article}

\usepackage[final]{acl}

\usepackage{times}
\usepackage{latexsym}

\usepackage[T1]{fontenc}

\usepackage[utf8]{inputenc}

\usepackage{microtype}

\usepackage{inconsolata}

\usepackage{graphicx}

\usepackage{times}
\usepackage{latexsym}
\usepackage{amssymb}
\usepackage{amsmath}
\usepackage{graphicx}
\usepackage{subcaption}
\usepackage{subfloat}
\usepackage{diagbox}
\usepackage{tikz}
\usepackage{colortbl}
\usepackage{xcolor}
\usepackage{multirow}
\usepackage{caption}
\usepackage{subcaption}
\usepackage{subfloat}
\usepackage{amsmath}
\usepackage{enumerate}
\usepackage{diagbox}
\usepackage{tikz}
\usepackage{colortbl}
\usepackage{xcolor}
\usepackage{multirow}
\usepackage{caption}
\usepackage{subcaption}
\usepackage{subfloat}
\usepackage{amsmath}
\usepackage{enumerate}
\usepackage{stfloats}
\usepackage[linesnumbered,ruled,vlined]{algorithm2e}
\usepackage{makecell}

\title{Instructions for *ACL Proceedings}

\title{Beyond Under-Alignment: Atomic Preference Enhanced Factuality Tuning for Large Language Models}

\author{Hongbang Yuan \textsuperscript{1,2},Yubo Chen \textsuperscript{1,2}, Pengfei Cao \textsuperscript{1,2},  Zhuoran Jin \textsuperscript{1,2},  \\ \textbf{Kang Liu \textsuperscript{1,2}, Jun Zhao \textsuperscript{1,2}} \\ \textsuperscript{1}The Laboratory of Cognition and Decision Intelligence for Complex Systems, \\ Institute of Automation, Chinese Academy of Sciences, Beijing, China  \\ \textsuperscript{2}School of Artificial Intelligence, University of Chinese Academy of Sciences, Beijing, China  \\
        \footnotesize{\texttt{\{hongbang.yuan, yubo.chen, pengfei.cao, zhuoran.jin, kliu, jzhao\} @nlpr.ia.ac.cn }} }

\begin{document}
\maketitle
\begin{abstract}

Large language models (LLMs) have achieved remarkable success but still tend to generate factually erroneous responses, a phenomenon known as hallucination. A recent trend is to use preference learning to fine-tune models to align with factuality. However, existing work primarily evaluates fine-tuned models on in-domain (ID) datasets and the factuality on out-of-domain (OOD) datasets remains underexplored. In this paper, we conduct a comprehensive evaluation of the factuality of different models tuned by various preference learning algorithms and demonstrate that their performance on OOD datasets either increases minimally or decreases. Subsequently, we reveal that the main cause of model's failure to uphold factuality under a distribution shift is \textbf{under-alignment}, rather than \textbf{over-alignment}, by analyzing the token distribution shift of the models before and after tuning. Finally, we propose \textbf{APEFT} (\textbf{A}tomic \textbf{P}reference \textbf{E}nhanced \textbf{F}actuality \textbf{T}uning), a framework that enhances model's awareness of factuality at the granularity of individual facts. Extensive experiments demonstrate that APEFT improves model performance by an average of $\boldsymbol{3.45\%}$ on both ID and OOD datasets, which is highly effective.

 
\end{abstract}

\section{Introduction}

Large language models (LLMs) have demonstrated surprising abilities \citep{chen2024chatgpts} and have achieved impressive advancements in many tasks \citep{bubeck2023sparks_of_agi}. However, they are still troubled by the issue of \textbf{hallucinations}, a phenomenon wherein LLMs generate seemingly convincing but factually erroneous responses \citep{zhang2023sirens_song,huang2023survey_of_hallucination}, which greatly hinders their deployment in practical scenarios \citep{cui2023chatlaw}.

Recently, some studies indicate that LLMs have the potential to avoid generating hallucinated answers. For example, LLMs mostly `know' the correct answer but fail to `tell' it \citep{li2023inferencetime,saunders2022selfcritiquing} and they show intrinsic uncertainty during hallucination occurrences \citep{manakul-etal-2023-selfcheckgpt,chen2024inside}. Therefore, a recent trend is to finetune LLMs to fully exploit their potential, typically using \textbf{preference learning} (or \textbf{preference tuning}) to steer the models towards desired behaviours and mitigate hallucinations \footnote{In our work, `mitigate hallucination' and `increase factuality' both refer to the same concept.} \citep{tian2023finetuning,lin2024flame}. Specifically, it involves collecting diverse model completions, annotating each with an overall preference score related to factuality, and refining models through the preference feedback. 

\begin{figure}
\centering

\includegraphics[width=\linewidth]{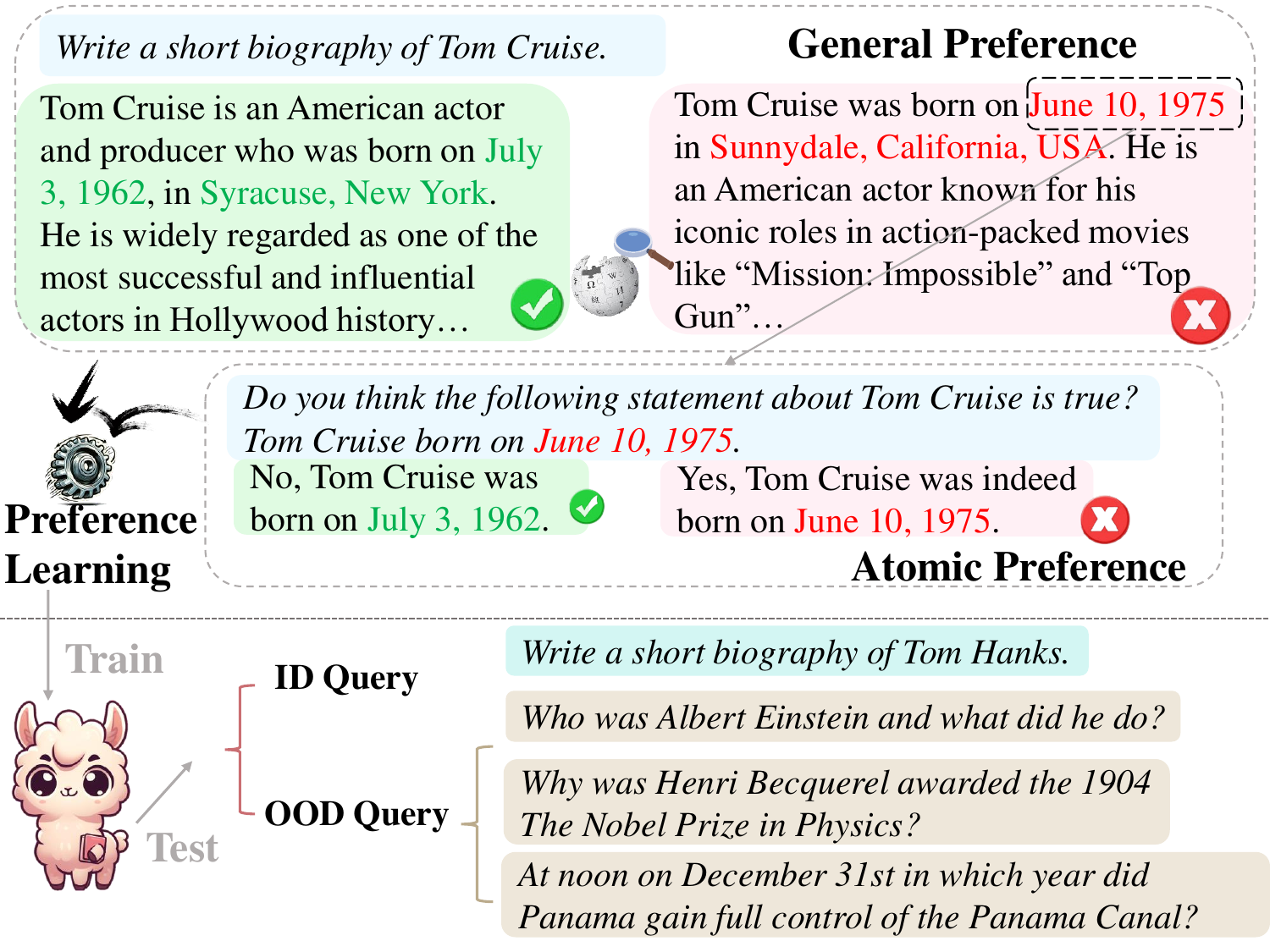}
\caption{Illustration of our work. The combination of general preferences and atomic preferences can enhance model factuality on OOD queries across various preference learning algorithms.}
\label{introduction}
\end{figure}

Despite being successful, existing work \citep{chen2024grath,tian2023finetuning} mainly uses test queries sourced from the same domain as the training dataset to evaluate the factuality of the tuned model, which is impractical in real-world applications. For example,  as shown in Figure \ref{introduction}, when training the model to be more factual in generating a biography of Tom Cruise, we expect performance improvements not only on in-domain (ID) queries, like generating a biography of Tom Hanks, but also on out-of-domain (OOD) queries, such as answering questions about Albert Einstein or the Panama Canal. Thus, extensive exploration of the factuality of the tuned models on OOD queries still remains incomplete.   

To analyze this generalization problem, we first conduct a comprehensive quantitative study on the OOD performance of the models tuned by various preference learning algorithms. In particular, the training data is constrained to the biography generation task and we train two LLMs on these data using five preference learning algorithms. Subsequently, we evaluate their performance on one ID dataset and OOD datasets. The OOD datasets incorporates various aspects related to model factuality, including generating factually accurate long-form responses consisting of several paragraphs, identifying false premises in user queries and answering with short responses on a scale of several phrases. In the end, the performance either increases minimally, or actually decreases, with the maximum drop in performance reaching $8.47\%$.



Therefore, we investigate the reasons for the ineffectiveness of the tuned models on OOD queries. We posit that there are two potential failure modes. One possibility is that the tuning process is particularly superficial, resulting in no significant change in the tuned model's behaviour in OOD settings (\textbf{under-alignment}). Alternatively, the tuning process may be adversely affected by spurious features in the training data, resulting in unintended changes in the tuned model's behaviour (\textbf{over-alignment}). For example, when asked about the contribution of Albert Einstein, the tuned models may produce nearly identical responses to those of their pre-tuned counterpart, or tend to generate vague and subjective sentences like `his insights are unparalleled', which is undesirable in this context.
To distinguish them, we measure the behaviour change of the model by directly comparing the shift in token distributions before and after tuning. We discover that the number of shifted tokens on OOD queries is less than a third of those on ID queries. \textit{It indicates that the primary cause of the ineffectiveness is under-alignment rather than over-alignment.}




Furthermore, we posit that under-alignment may result from the the original factual preference feedback, which is predominantly based on paragraphs and does not adequately inform the model about the factuality of individual facts. Therefore, we propose \textbf{A}tomic \textbf{P}reference \textbf{E}nhanced \textbf{F}actuality \textbf{T}uning (APEFT), a framework that enhances model's awareness of factuality at the granularity of individual facts. Initially, we define the preferences collected on one specific task as \textbf{general preferences}. Firstly, we break the responses in general preferences into single sentences containing only one piece of knowledge. Subsequently, we assess the extent to which the model knows these pieces of knowledge by collecting stochastically sampled responses to a knowledge detection prompt. Then, we select the contradicted responses along with the knowledge detection prompt to construct preference pairs, denoted as \textbf{atomic preferences}. Finally, we train the models based on both the atomic preferences and the general preferences. Experimental results demonstrate performance improvements across almost all the preference learning algorithms.
Additionally, we briefly explore whether merely increasing the quantity and improving the quality of training preference pairs enhances factuality and find that they do not necessarily lead to performance gains.


Our contributions can be summarized as follows:
\begin{itemize}
    \item We conduct a comprehensive evaluation of the factuality of different models tuned by various preference learning algorithms. We demonstrate that the model's performance on OOD queries either increases minimally or even decreases after the tuning.

    \item We provide two potential failure modes explaining why the models fail to uphold their factuality under a distribution shift: under-alignment and over-alignment. We reveal that under-alignment is the primary cause by analyzing the differences in token distributions of the models before and after tuning.

    \item We propose \textbf{APEFT} (\textbf{A}tomic \textbf{P}reference \textbf{E}nhanced \textbf{F}actuality \textbf{T}uning), a framework that enhances model's awareness of factuality at the granularity of individual facts. Extensive experiments demonstrate its effectiveness and the model performance increases an average of $\boldsymbol{3.45\%}$ on both the OOD and ID datasets. \footnote{The code and datasets will be available after acceptance.}
\end{itemize}

\section{Related Work}  

\paragraph{Hallucination} 
Hallucinations greatly impede the practical applications of LLMs \citep{survey_of_hallucination_in_VLLMs,mündler2023selfcontradictory} and many studies have been conducted to address the issue \citep{trivedi-etal-2023-interleaving-retrieval,gou2023critic}. Since LLMs have shown internal awareness when hallucination occurs \citep{azaria-mitchell-2023-internal,manakul-etal-2023-selfcheckgpt}, some methods propose to intervene the internal representations during inference to mitigate hallucinations \citep{li2023inferencetime,chuang2023dola,zou2023transparency} while others directly fine-tune the LLMs to be more factual in the first place \citep{tian2023finetuning,chen2024grath,lin2024flame}. 
However, they primarily use test data sourced from the same domain as the training data to evaluate factuality, and fail to consider the generalization of their methods on out-of-domain datasets.

\paragraph{Alignment}
Aligning LLMs with human preferences is a crucial step to steer already-capable models towards desired behaviors \citep{burns2023weaktostrong,liu2024makes}. It typically involves instruction tuning and reinforcement learning from human feedback \citep{DBLP:conf/iclr/SanhWRBSACSRDBX22,ouyang2022training,bai2022training,less_is_more}. Since relative preference scores are more feasible than expert demonstrations \citep{DBLP:conf/nips/RafailovSMMEF23}, many studies develop methods to fine-tune models with preferences \citep{azar2023ipo,ethayarajh2024kto,xu2024cpo}. However, as human preferences in many datasets tend to under-represent the factuality of the responses \citep{hosking2024human}, it's still unclear how to construct preference pairs that can enhance the factuality of LLMs.




\section{A Comprehensive Evaluation}
\label{general_preference_evaluation}
In this section, we conduct comprehensive quantitative study on the performance of models tuned by various preference learning algorithms. We demonstrate that nearly all the preference learning algorithms deliver unsatisfactory performance on the OOD datasets. 



\subsection{Data Construction}
Initially, we construct a preference dataset that focuses on the factuality of the responses.

Formally, we prompt a pre-trained LLM $\pi_\theta$ with a prompt $x$ to produce pairs of answers: $(y_1,y_2)\sim\pi_\theta(y|x)$. Then we compare which response aligns more closely with external knowledge sources. The preferred response with fewer factual errors is denoted as $y_w$, while the dispreferred response with more factual errors is denoted as $y_l$. 

\urlstyle{same}
Specifically, we prompt the LLMs to generate biographies of famous people, with names sampled from a list of the most popular individuals \footnote{\url{https://today.yougov.com/ratings/entertainment/popularity/people/all}}. To measure the factuality of the responses, we employ FActScore \citep{min-etal-2023-factscore}, which computes the percentage of individual facts in the generated responses that are supported by the knowledge source. For each prompt, we generate N responses, create $\binom{N}{2}$ preference pairs, and then filter out those pairs that have identical factuality scores. Finally, we obtain a dataset of $N$ preference pairs $D=\{x^i,y_w^i,y_l^i\}_{i=1}^{N}$.

We employ LLaMA-2-7B-Chat \citep{touvron2023llama} and LLaMA-3-8B-Instruct \footnote{\url{https://huggingface.co/meta-llama/Meta-Llama-3-8B-Instruct}} as the base model of our experiments. Ultimately, we construct 2777 preference pairs for LLaMA-3-8B-Instruct and 2730 preference pairs for LLaMA-2-7B-Chat. 


\subsection{Training}
We primarily use reward-free preference learning algorithms to avoid the complex process of reinforcement learning without the loss of generality. Specifically, we employ the following representative methods:

(1) \textbf{DPO }\citep{DBLP:conf/nips/RafailovSMMEF23}, which directly optimizes a simple classification loss function without the explicit usage of reward model.

(2) \textbf{RSO }\citep{liu2024statistical}, which proposes to employ a hinge loss to replace the sigmoid loss in the classification loss function in DPO. 

(3) \textbf{IPO} \citep{DBLP:conf/aistats/AzarGPMRVC24}, which designs a new general objective to avoid over-fitting to the preference dataset.

(4) \textbf{KTO} \citep{ethayarajh2024kto}, which directly maximizes the utility of generated responses instead of the log-likelihood of preferences.

(5) \textbf{CPO} \citep{xu2024contrastive}, which derives an approximation of the DPO loss that is more memory-efficient and speed-efficient. 

Further details about these methods are provided in Appendix \ref{sec:preference_learning_algorithms}.

\urlstyle{same}
For the training hyperparameters, we set training epochs to $3$, learning rate to $1e^{-6}$, batch size to $4$ and gradient accumulation step to $4$. We use LLaMA-Factory \footnote{\url{https://github.com/hiyouga/LLaMA-Factory}} to train the models using full parameter fine-tuning. All of our experiments are conducted on 4 Nvidia A100-80G GPUs.

\begin{table*}[]
\centering
\resizebox{\linewidth}{!}{
\begin{tabular}{ccccc}
\hline
Dataset & Samples & Example & Task  & Type 
\\ \arrayrulecolor{gray} \hline

Bio     & 44      & Write a short   biography of Robert Duvall.   & Biography Generation                                                                      & ID   \\
FAVA    & 100     & Explain Hypermarcas, including information about industry, key person.   & Open-ended Generation                                         & OOD  \\
FPQA    & 986     & Why was Albert Einstein awarded the 1922 The Nobel Prize in Physics? & False Premise QA     & OOD  \\
KUQA    & 946     & What languages did Mila Kunis speak as a child? & Knowledge-based QA  & OOD 
\\ 
\arrayrulecolor{black}
\hline
\end{tabular}
}
\caption{Statistics and examples of the evaluation datasets.}
\label{dataset_statistics}
\end{table*}

\subsection{Evaluation}

We evaluate the factuality of the tuned models from two aspects: in-domain and out-of-domain. For the in-domain evaluation, we use the same biography generation task as the training data. For the out-of-domain evaluation, we choose the following datasets:

(1) \textbf{FAVA} \citep{mishra2024finegrained}, which contains information-seeking  queries on open-ended topics, aims to benchmark model's long-form factuality in a variety of domains.

(2) \textbf{FPQA} \citep{yuan2024whispers}, which contains questions that have false premises, aims to evaluate the model's factuality when faced with such unanswerable questions.

(3) \textbf{KUQA} \citep{jin2024rwku}, which contains questions about particular entities and their neighbours, aims to assess the model's knowledge through its ability to generate short-form phrases.

To facilitate evaluation, we select a subset of each dataset. The statistics and concrete examples of each dataset are shown in Table \ref{dataset_statistics}. 

\begin{figure}[t]
\centering
\includegraphics[width=\linewidth]
{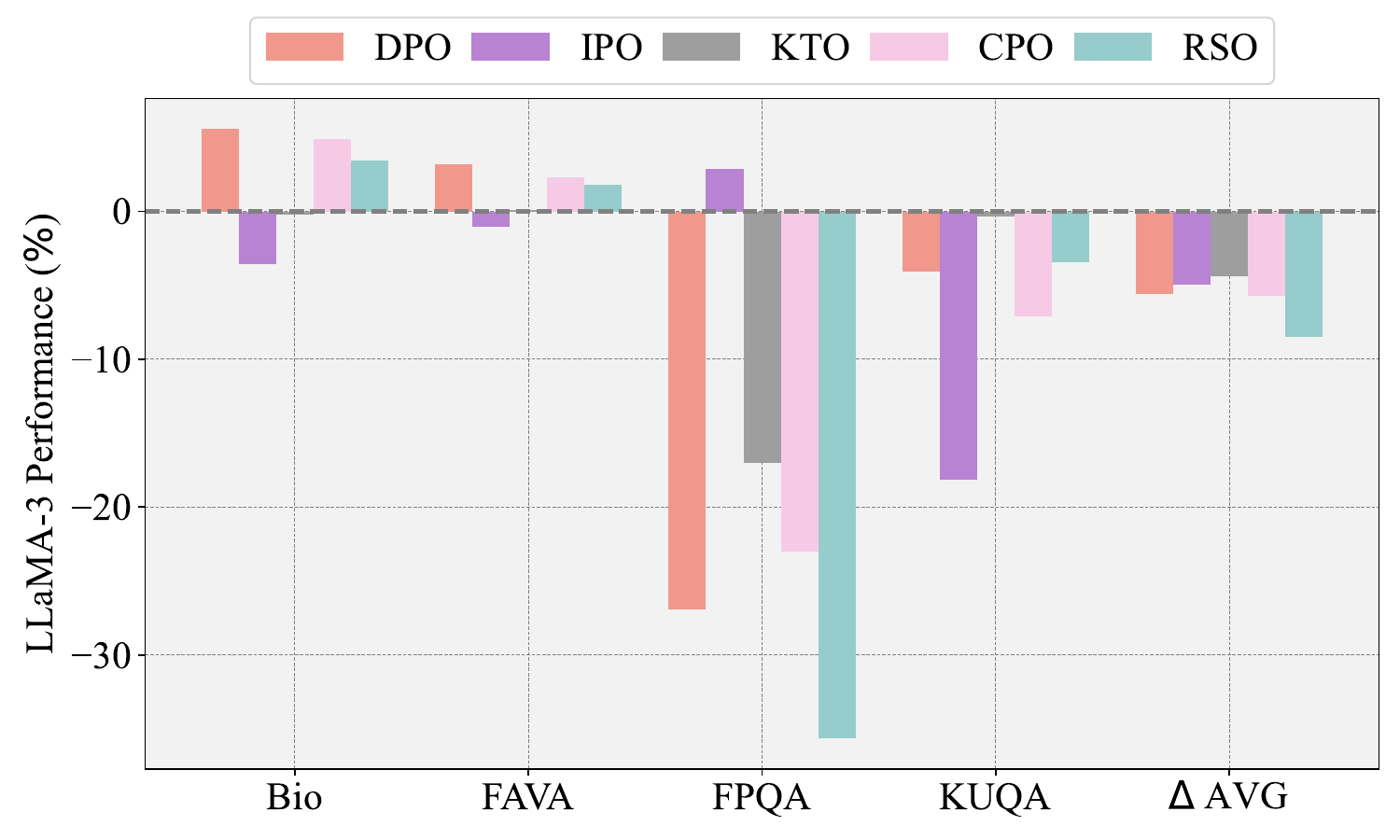}
\caption{Performance changes of LLaMA-3-8B-Instruct before and after tuning using various preference learning techniques.}
\label{pilot_study}
\end{figure}


\subsection{Results and Analysis}
We train the models using various preference learning algorithms using our constructed preference pairs. Then, we depict the performance changes of LLaMA-3-8B-Instruct before and after tuning in Figure \ref{pilot_study}. Additional results are presented in Appendix \ref{sec:llama2_general_performance}. We can draw the following observations: \textbf{(1)} Our constructed preference pairs can improve model factuality on in-domain datasets. For example, apart from the IPO, nearly all the preference learning algorithms demonstrate an improvement in the FActScore on Bio dataset. \textbf{(2)} Nearly all the preference learning algorithms perform poorly on OOD datasets. For example, the performance on FAVA increases minimally while the performance actually decreases on FPQA and KUQA.
\section{Under-Aligned or Over-Aligned?}
Given the unsatisfactory performance of the tuned models, we set out to better understand the reasons for the ineffectiveness. We start by proposing two potential failure modes, namely under-alignment and over-alignment, and then we conduct the token distribution analysis proposed by \citet{lin2023unlocking} to empirically determine the primary cause.


\subsection{Two Failure Modes}
Intuitively, the failure may be caused either by insufficient learning from the preference pairs or by excessive attention to spurious features that are unrelated to factuality. We termed these two cases under-alignment and over-alignment. We describe them in detail in the following paragraphs:  

\paragraph{Under-Alignment} The preference learning process is particularly superficial, resulting in no significant change in the tuned model's behaviour in OOD settings. For example, the tuned models may produce the same responses as their pre-tuned counterpart because they didn't fully aware of the factuality conveyed by the training preference pairs.

\paragraph{Over-Alignment} The preference learning process is adversely affected by spurious features in the training data, resulting in unintended changes in model behaviour in OOD settings. For example, the model tends to generate vague and subjective queries after training on the biography generation task, thus failing to answer questions such as the specific contributions of Albert Einstein or when Panama gained full control of the Panama Canal.

Notably, it is significant to distinguish them as it influences how we proceed to address the generalization failure. If the under-alignment is the primary cause, enhancing the model's awareness of factuality becomes necessary. Conversely, if the over-alignment is the primary cause, it would be imperative to revise the learning process or the current preference pairs to minimize the impact of spurious features.

\subsection{Empirical Validation}
To quantitatively measure the behaviour change during the preference learning, we directly compare the differences in token distribution of the models before and after tuning.

\paragraph{Method} Intuitively, if the model's behaviour changes significantly after the fine-tuning process, the predicted tokens using the same context will also change dramatically. Specifically, the response of the fine-tuned model is denoted as $\{t_1,t_2,...,t_N\}$ and at each token position, the next token prediction probability distribution is denoted as $P_{aligned}$. For each token position $t_i$, the context $\{t_1,t_2,...,t_{i-1}\}$ is input again to the pre-tuned counterpart and the resulting distribution is denoted as $P_{base}$. We measure how the rank of $t_i$ in $P_{aligned}$ changes compared to its rank in $P_{base}$ and define the difference as $\delta$. The greater the difference, the more significant the change in model behaviour after the tuning. We refer to a tokens as `shifted token' if its rank difference is greater than zero.


\paragraph{Results and Analysis}

\begin{figure}[t]
\centering
\subfloat[Bio Dataset]{\includegraphics[width=\linewidth]{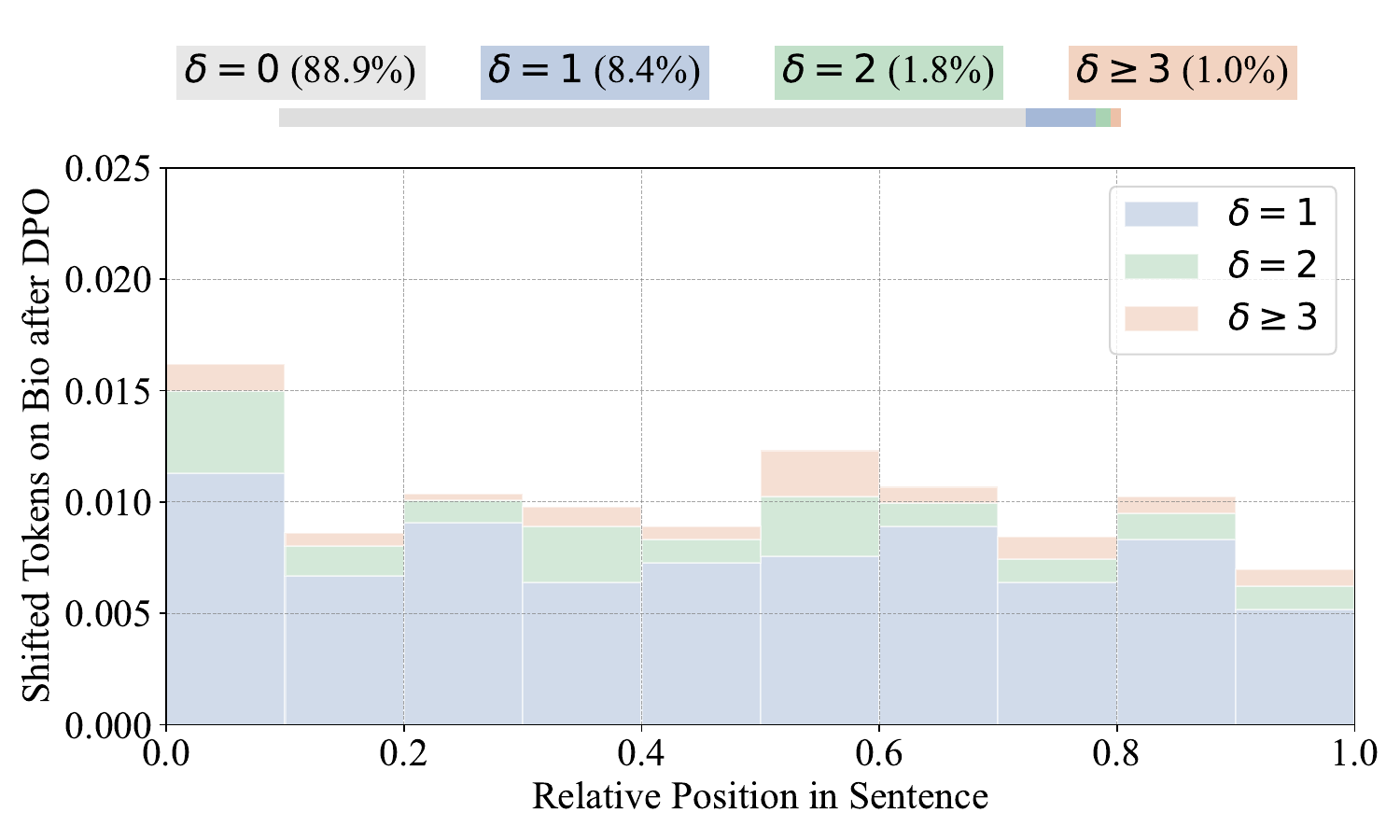}\label{attn_head_influence}} \hfill
\subfloat[FAVA Dataset]{\includegraphics[width=\linewidth]{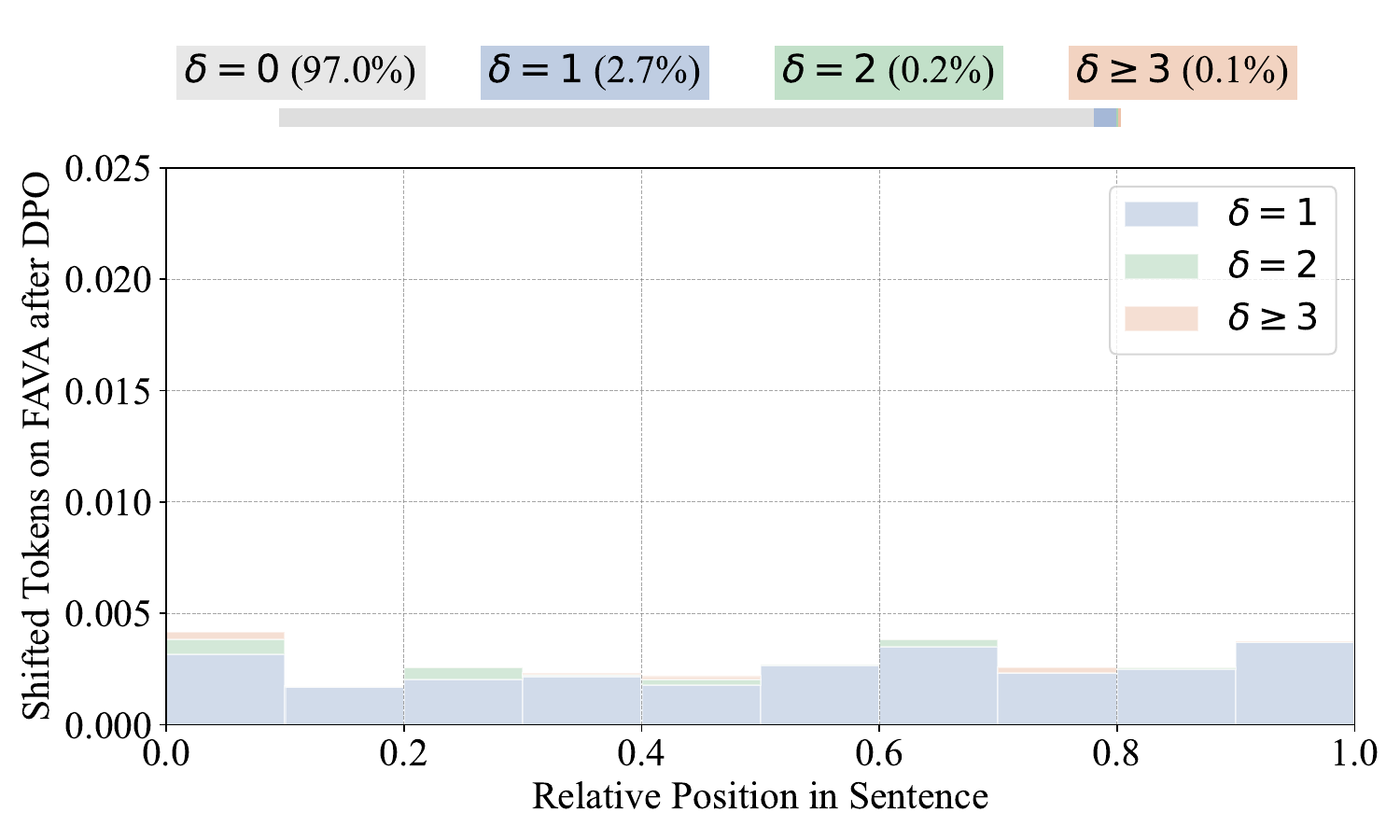}\label{attn_head_influence}} 
\caption{Token shift analysis on LLaMA-2 trained by DPO. More results are presented in Appendix \ref{sec:token_shift_analysis}.}
\label{token_shift_result}
\end{figure}

We conduct experiments on the ID dataset Bio and the OOD dataset FAVA on LLaMA-2-7B-Chat trained by various learning algorithms. We calculate the frequency of various levels of the shifted tokens along with their relative positions in each sentence. The experimental results are shown in Figure \ref{token_shift_result}. We draw the following observation: shifted tokens are more prevalent in the in-domain dataset than in the out-of-domain dataset. The frequency bars in the Bio figure are almost three times as high as those in the Fava figure. This indicates that the the model's behaviour changes far less on OOD queries than on ID queries and the performance drop on OOD queries is more likely a result of under-alignment rather than over-alignment. Therefore, it's essential to design methods to enhance model's awareness of factuality.


\section{Atomic Preference Enhanced Factuality Tuning}

In this section, we introduce APEFT, a framework designed to enhance the model's awareness of factuality at the granularity of individual facts. We start with a detailed description of our framework and then demonstrate its effectiveness by extensive quantitative experiments.

\subsection{Framework Description}
\label{framework_description}

\begin{figure}[t]
\centering
\includegraphics[width=\linewidth]{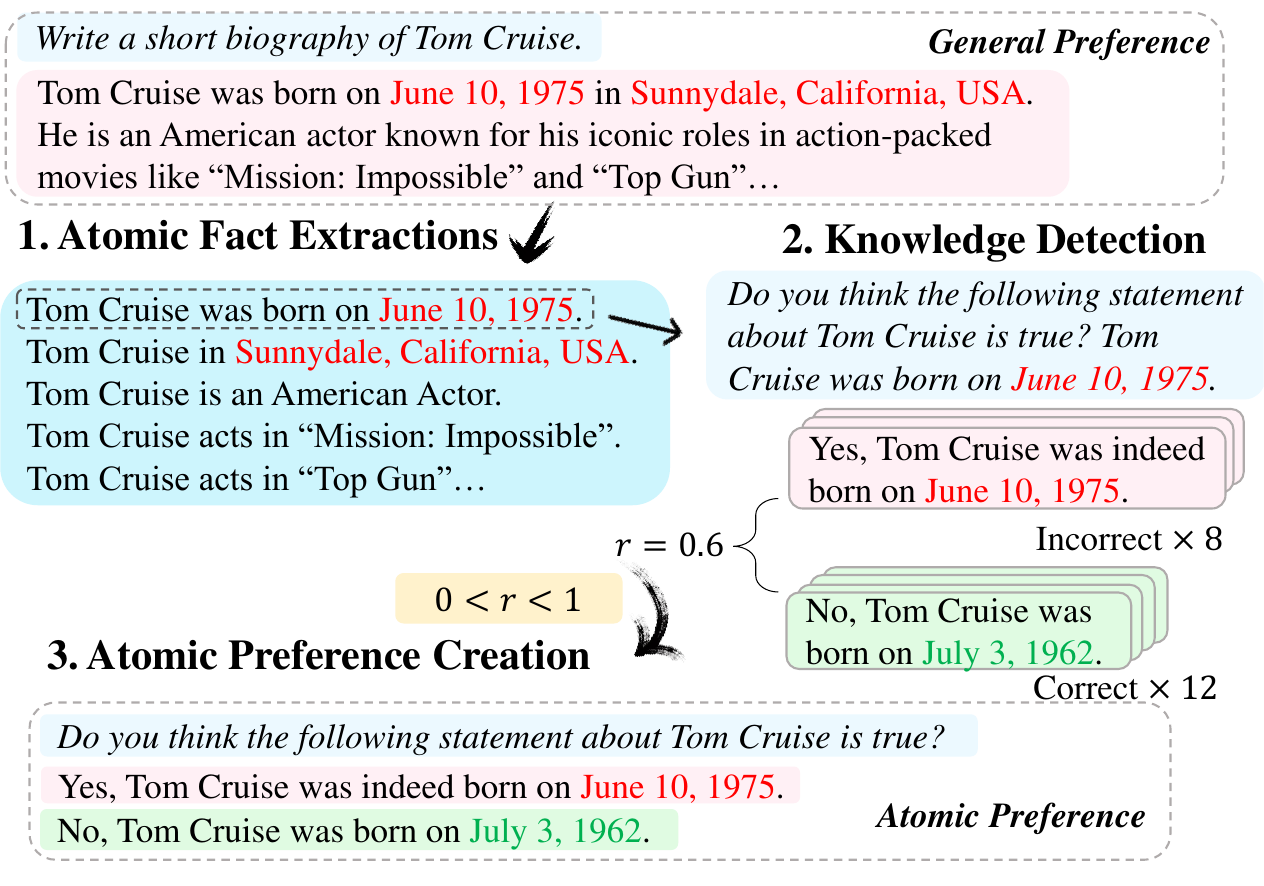}
\caption{Illustration of our proposed APEFT. It creates atomic preferences from the responses in original general preferences to enhance model's awareness of factuality at the granularity of individual facts.}
\label{method}
\end{figure}

We posit that the original factual preference feedback, which is mainly based on paragraphs, is too coarse-grained and insufficient to convey the factuality of individual facts. For example, as shown in Figure \ref{introduction}, the general preferences contain too much information, such as the date and place of Tom Cruise’s birth, the films in which he starred. Consequently, the model is unable to discern exactly which fact is incorrect, leading to under-alignment on OOD queries. Thus it's essential to construct preferences at the granularity of each individual fact. In particular, as shown in Figure \ref{method}, our method APEFT is divided into three steps: atomic fact extraction, knowledge detection and atomic preference creation. We will describe each step in detail sequentially.

\paragraph{Atomic Fact Extraction}
Firstly, we extract individual atomic facts from the general preference dataset. Following \citep{min-etal-2023-factscore,wei2024longform}, we define the atomic facts as short sentences containing only one piece of information.  We extract atomic facts of both the preferred response $y_w$ and the dispreferred response $y_l$ for every preference pair by prompting ChatGPT.

\paragraph{Knowledge Detection}
Subsequently, we assess the extent to which model knows the knowledge contained in each atomic fact. We ask the model to decide whether the provided atomic fact is true or false and detect the model's internal knowledge through the percentage of correct responses among the stochastically sampled responses. 

We employ multinomial sampling as the decoding strategy and set temperature to 1 to increase randomness and diversity. With the percentage of the correct responses after the sampling denoted as $r$, the \textbf{unknown}, \textbf{potentially-known} and \textbf{known} atomic facts are defined as when ($r=0$), ($0<r<1$) and ($r=1$) respectively. We propose that our focus should be on improving those potentially-known atomic facts.

\paragraph{Atomic Preference Creation}
Finally, we construct atomic preferences from previously identified potentially-known atomic facts. We direct use the knowledge detection prompt as the user query $x$ and the two contradictory responses as preference $(y_w,y_l)$. The atomic preferences are mixed together with the general preferences to enhance the factuality of LLMs. With the original general preferences denoted as $D_g$ and the newly constructed atomic preferences denoted as $D_a$, we aim to optimize the following generic objective function:  

\begin{small}
    \begin{equation*}
O\left(\pi_\theta\right)=\underset{\left(\mathbf{y}, x\right) \sim D_g \cup D_a}{\mathbb{E}} \left[r(x, \mathbf{y})-\beta \log \frac{\pi_\theta(\mathbf{y} \mid x)}{\pi_{\mathrm{ref}}(\mathbf{y} \mid x)}\right]
\end{equation*}
\end{small}
where the data $(x,\mathbf{y})$ can either be $(x,y_w)$ or $(x,y_l)$, $r$ is the reward model and $\beta$ is a regularisation term controlling the Kullback-Leibler divergence between the policy model $\pi_\theta$ and the reference model  $\pi_{\mathrm{ref}}$. Further details about the objective function are presented Appendix \ref{sec:preference_learning_algorithms}.

\begin{table*}[t!]
\centering
\resizebox{0.83\linewidth}{!}{
\begin{tabular}{c|ccc|ccc|c|c|l}
\hline
\arrayrulecolor{gray!100} 
\multirow{2}{*}{Methods} & \multicolumn{3}{c|}{\textbf{Bio}} & \multicolumn{3}{c|}{\textbf{FAVA}} & \textbf{FPQA}                & \textbf{KUQA} & \multicolumn{1}{c}{\multirow{2}{*}{\textbf{AVG(↑)}}} \\
                                  & FS(\%)(↑)    & NC(↑)   & NE(↓)   & FS(\%)(↑)    & NC(↑)    & NE(↓)   & Acc(↑)                       & Acc(↑)        & \multicolumn{1}{c}{}   \\  \hline      \multicolumn{10}{c}{\textit{LLaMA-2-7B-Chat}} \\ \hline
\multicolumn{1}{l|}{Vanilla} &  82.11   &  27.41   &  6.02   &  60.37   &  18.46   &  11.96   &  22.99   &  68.09   &  58.39  \\ 
  \hline \multicolumn{1}{l|}{+DPO} &   \cellcolor[HTML]{c2ddec}87.35   &   \cellcolor[HTML]{d5e7f1}29.23   &   \cellcolor[HTML]{d7e8f2}4.30   &   \cellcolor[HTML]{e7f2f7}62.37   &   \cellcolor[HTML]{eff6fa}19.14   &   \cellcolor[HTML]{f5f9fb}11.53   &   \cellcolor[HTML]{b3d4e7}36.18   &   \cellcolor[HTML]{fedccd}65.11   &   \cellcolor[HTML]{cce3ef}62.75  \\ 
 \textit{w/rand} &   \cellcolor[HTML]{c1dceb}87.45   &   \cellcolor[HTML]{d0e4f0}29.45   &   \cellcolor[HTML]{d5e8f2}4.23   &   \cellcolor[HTML]{cde3ef}64.66   &   \cellcolor[HTML]{dcebf4}19.95   &   \cellcolor[HTML]{e5f0f6}10.83   &   \cellcolor[HTML]{dbebf3}29.22   &   \cellcolor[HTML]{d6e8f2}71.62   &   \cellcolor[HTML]{c7dfed}63.24  \\ 
 \textit{w/atom} &   \cellcolor[HTML]{bad8e9}88.08   &   \cellcolor[HTML]{ecf4f9}28.23   &   \cellcolor[HTML]{cde3ef}3.86   &   \cellcolor[HTML]{d9e9f3}63.66   &   \cellcolor[HTML]{f4f9fb}18.90   &   \cellcolor[HTML]{eaf3f8}11.08   &   \cellcolor[HTML]{a2cbe2}39.03   &   \cellcolor[HTML]{eef5f9}69.56   &   \cellcolor[HTML]{b2d4e6}\textbf{65.08}  \\ 
  \hline \multicolumn{1}{l|}{+IPO} &   \cellcolor[HTML]{d6e8f2}85.66   &   \cellcolor[HTML]{d9eaf3}29.05   &   \cellcolor[HTML]{e8f2f8}5.05   &   \cellcolor[HTML]{d1e5f0}64.30   &   \cellcolor[HTML]{e5f0f7}19.56   &   \cellcolor[HTML]{f0f6fa}11.32   &   \cellcolor[HTML]{aed2e5}37.03   &   \cellcolor[HTML]{fdd7c6}64.67   &   \cellcolor[HTML]{cae2ee}62.92  \\ 
 \textit{w/rand} &   \cellcolor[HTML]{d7e8f2}85.55   &   \cellcolor[HTML]{dfedf5}28.77   &   \cellcolor[HTML]{e2eff6}4.77   &   \cellcolor[HTML]{ddecf4}63.25   &   \cellcolor[HTML]{eff6fa}19.15   &   \cellcolor[HTML]{f0f6fa}11.32   &   \cellcolor[HTML]{fdbb9d}11.29   &   \cellcolor[HTML]{deecf4}70.93   &   \cellcolor[HTML]{fef7f4}57.75  \\ 
 \textit{w/atom} &   \cellcolor[HTML]{b9d8e9}88.17   &   \cellcolor[HTML]{fdbfa3}24.66   &   \cellcolor[HTML]{bfdbeb}3.25   &   \cellcolor[HTML]{c0dceb}65.84   &   \cellcolor[HTML]{deecf4}19.88   &   \cellcolor[HTML]{e3eff6}10.75   &   \cellcolor[HTML]{d3e7f1}30.49   &   \cellcolor[HTML]{cbe2ee}72.58   &   \cellcolor[HTML]{bbd9e9}\textbf{64.27}  \\ 
  \hline \multicolumn{1}{l|}{+KTO} &   \cellcolor[HTML]{f6fafc}82.88   &   \cellcolor[HTML]{eef5f9}28.14   &   \cellcolor[HTML]{f8fbfd}5.75   &   \cellcolor[HTML]{f4f9fb}61.25   &   \cellcolor[HTML]{fbfcfd}18.63   &   \cellcolor[HTML]{fefbf9}12.12   &   \cellcolor[HTML]{8cbfdb}43.99   &   \cellcolor[HTML]{fcac87}60.92   &   \cellcolor[HTML]{d2e6f1}62.26  \\ 
 \textit{w/rand} &   \cellcolor[HTML]{d9eaf3}85.34   &   \cellcolor[HTML]{eef5f9}28.14   &   \cellcolor[HTML]{e4f0f6}4.86   &   \cellcolor[HTML]{fafcfd}60.79   &   \cellcolor[HTML]{fcfdfe}18.56   &   \cellcolor[HTML]{fef1ec}12.53   &   \cellcolor[HTML]{fef8f5}21.84   &   \cellcolor[HTML]{e5f0f7}70.30   &   \cellcolor[HTML]{f1f7fa}59.57  \\ 
 \textit{w/atom} &   \cellcolor[HTML]{e2eef5}84.62   &   \cellcolor[HTML]{e7f1f7}28.43   &   \cellcolor[HTML]{ecf4f9}5.20   &   \cellcolor[HTML]{d5e7f2}63.98   &   \cellcolor[HTML]{d8e9f3}20.12   &   \cellcolor[HTML]{eff6fa}11.27   &   \cellcolor[HTML]{9ac7df}40.40   &   \cellcolor[HTML]{fca781}60.52   &   \cellcolor[HTML]{d1e5f0}\textbf{62.38}  \\ 
  \hline \multicolumn{1}{l|}{+CPO} &   \cellcolor[HTML]{aed1e5}89.15   &   \cellcolor[HTML]{e6f1f7}28.48   &   \cellcolor[HTML]{c5dfed}3.52   &   \cellcolor[HTML]{e8f2f8}62.31   &   \cellcolor[HTML]{e4f0f6}19.61   &   \cellcolor[HTML]{fefefe}11.97   &   \cellcolor[HTML]{bcdaea}34.49   &   \cellcolor[HTML]{fef7f4}67.44   &   \cellcolor[HTML]{c5dfed}63.35  \\ 
 \textit{w/rand} &   \cellcolor[HTML]{aad0e4}89.42   &   \cellcolor[HTML]{cce3ef}29.59   &   \cellcolor[HTML]{c6dfed}3.55   &   \cellcolor[HTML]{cae1ee}64.92   &   \cellcolor[HTML]{ddecf4}19.93   &   \cellcolor[HTML]{e9f2f8}11.02   &   \cellcolor[HTML]{d2e6f1}30.70   &   \cellcolor[HTML]{c7e0ed}72.92   &   \cellcolor[HTML]{b8d7e9}\textbf{64.49}  \\ 
 \textit{w/atom} &   \cellcolor[HTML]{a9cfe4}89.54   &   \cellcolor[HTML]{c5dfed}29.89   &   \cellcolor[HTML]{c5deec}3.50   &   \cellcolor[HTML]{bfdbeb}65.91   &   \cellcolor[HTML]{d9eaf3}20.07   &   \cellcolor[HTML]{d9e9f3}10.31   &   \cellcolor[HTML]{d3e6f1}30.59   &   \cellcolor[HTML]{dfedf5}70.81   &   \cellcolor[HTML]{bcd9ea}64.21  \\ 
  \hline \multicolumn{1}{l|}{+RSO} &   \cellcolor[HTML]{bfdbeb}87.63   &   \cellcolor[HTML]{eaf3f8}28.30   &   \cellcolor[HTML]{d1e5f0}4.02   &   \cellcolor[HTML]{eaf3f8}62.18   &   \cellcolor[HTML]{edf5f9}19.22   &   \cellcolor[HTML]{f8fbfd}11.69   &   \cellcolor[HTML]{b5d6e8}35.76   &   \cellcolor[HTML]{fedccc}65.08   &   \cellcolor[HTML]{cde3ef}62.66  \\ 
 \textit{w/rand} &   \cellcolor[HTML]{bfdbeb}87.59   &   \cellcolor[HTML]{bad8e9}30.41   &   \cellcolor[HTML]{d7e8f2}4.30   &   \cellcolor[HTML]{deecf4}63.23   &   \cellcolor[HTML]{ebf3f8}19.32   &   \cellcolor[HTML]{f5f9fc}11.56   &   \cellcolor[HTML]{fefaf8}22.15   &   \cellcolor[HTML]{c4deec}73.18   &   \cellcolor[HTML]{daeaf3}61.54  \\ 
 \textit{w/atom} &   \cellcolor[HTML]{afd2e5}89.06   &   \cellcolor[HTML]{bfdbeb}30.16   &   \cellcolor[HTML]{c6dfed}3.57   &   \cellcolor[HTML]{c8e0ed}65.11   &   \cellcolor[HTML]{d9eaf3}20.07   &   \cellcolor[HTML]{e5f0f6}10.83   &   \cellcolor[HTML]{a2cbe1}39.13   &   \cellcolor[HTML]{fef8f6}67.57   &   \cellcolor[HTML]{b0d3e6}\textbf{65.22 }

\\ \arrayrulecolor{black} \hline
\arrayrulecolor{gray}

\multicolumn{10}{c}{\textit{LLaMA-3-8B-Instruct}} \\ \hline

\multicolumn{1}{l|}{Vanilla} &  86.37   &  33.86   &  5.43   &  69.14   &  26.95   &  11.97   &  82.91   &  74.29   &  78.18  \\ 
  \hline \multicolumn{1}{l|}{+DPO} &   \cellcolor[HTML]{bedbea}91.94   &   \cellcolor[HTML]{d3e6f1}35.77   &   \cellcolor[HTML]{cbe2ee}3.20   &   \cellcolor[HTML]{daeaf3}72.32   &   \cellcolor[HTML]{cbe2ee}29.19   &   \cellcolor[HTML]{e3eff6}10.78   &   \cellcolor[HTML]{fc8c59}56.01   &   \cellcolor[HTML]{fdd0bb}70.21   &   \cellcolor[HTML]{fdbfa2}72.62  \\ 
 \textit{w/rand} &   \cellcolor[HTML]{c0dceb}91.79   &   \cellcolor[HTML]{bfdbeb}36.61   &   \cellcolor[HTML]{cde3ef}3.27   &   \cellcolor[HTML]{e1eef5}71.68   &   \cellcolor[HTML]{fdfefe}27.02   &   \cellcolor[HTML]{d9e9f3}10.32   &   \cellcolor[HTML]{fc8c59}50.21   &   \cellcolor[HTML]{d5e7f1}77.90   &   \cellcolor[HTML]{fdc2a7}72.90  \\ 
 \textit{w/atom} &   \cellcolor[HTML]{cde3ef}90.69   &   \cellcolor[HTML]{e1eef5}35.16   &   \cellcolor[HTML]{d5e8f2}3.64   &   \cellcolor[HTML]{fafcfd}69.49   &   \cellcolor[HTML]{fedaca}25.36   &   \cellcolor[HTML]{ddecf4}10.53   &   \cellcolor[HTML]{fca57d}67.30   &   \cellcolor[HTML]{dcebf4}77.27   &   \cellcolor[HTML]{fee8dd}\textbf{76.19}  \\ 
  \hline \multicolumn{1}{l|}{+IPO} &   \cellcolor[HTML]{fdd6c3}82.81   &   \cellcolor[HTML]{fef6f3}33.50   &   \cellcolor[HTML]{fed8c7}7.09   &   \cellcolor[HTML]{fef3ed}68.11   &   \cellcolor[HTML]{fefbf9}26.78   &   \cellcolor[HTML]{fefefe}11.93   &   \cellcolor[HTML]{eef5f9}85.76   &   \cellcolor[HTML]{fc8c59}56.15   &   \cellcolor[HTML]{fdc5ac}73.21  \\ 
 \textit{w/rand} &   \cellcolor[HTML]{fef5f1}85.55   &   \cellcolor[HTML]{fc8c59}28.77   &   \cellcolor[HTML]{eff6fa}4.77   &   \cellcolor[HTML]{feeae0}67.32   &   \cellcolor[HTML]{fdc0a5}24.25   &   \cellcolor[HTML]{f8fbfc}11.68   &   \cellcolor[HTML]{fef5f0}81.22   &   \cellcolor[HTML]{b0d3e6}81.11   &   \cellcolor[HTML]{f7fbfc}78.80  \\ 
 \textit{w/atom} &   \cellcolor[HTML]{feece4}84.75   &   \cellcolor[HTML]{fbfcfd}34.02   &   \cellcolor[HTML]{feeee6}6.16   &   \cellcolor[HTML]{fdd2be}65.27   &   \cellcolor[HTML]{fdc9b1}24.61   &   \cellcolor[HTML]{fef8f6}12.24   &   \cellcolor[HTML]{b3d5e7}95.99   &   \cellcolor[HTML]{feeae1}72.53   &   \cellcolor[HTML]{eef5f9}\textbf{79.63 } \\ 
  \hline \multicolumn{1}{l|}{+KTO} &   \cellcolor[HTML]{fefcfb}86.17   &   \cellcolor[HTML]{fee4d8}32.70   &   \cellcolor[HTML]{fcfdfe}5.32   &   \cellcolor[HTML]{fdfefe}69.24   &   \cellcolor[HTML]{f8fbfd}27.22   &   \cellcolor[HTML]{fefcfa}12.10   &   \cellcolor[HTML]{fc9d72}65.93   &   \cellcolor[HTML]{fefbf9}73.95   &   \cellcolor[HTML]{fdccb6}73.82  \\ 
 \textit{w/rand} &   \cellcolor[HTML]{feefe9}85.06   &   \cellcolor[HTML]{feebe2}33.00   &   \cellcolor[HTML]{fef6f2}5.82   &   \cellcolor[HTML]{fdd3c0}65.35   &   \cellcolor[HTML]{fdbb9d}24.02   &   \cellcolor[HTML]{eaf3f8}11.10   &   \cellcolor[HTML]{fc8c59}48.73   &   \cellcolor[HTML]{a1cae1}82.44   &   \cellcolor[HTML]{fca57d}70.40  \\ 
 \textit{w/atom} &   \cellcolor[HTML]{feede5}84.86   &   \cellcolor[HTML]{fef4f0}33.41   &   \cellcolor[HTML]{fef1eb}6.02   &   \cellcolor[HTML]{fdccb5}64.73   &   \cellcolor[HTML]{fdc3a9}24.37   &   \cellcolor[HTML]{d8e9f2}10.29   &   \cellcolor[HTML]{afd2e6}96.80   &   \cellcolor[HTML]{dbebf3}77.35   &   \cellcolor[HTML]{dfedf5}\textbf{80.93}  \\ 
  \hline \multicolumn{1}{l|}{+CPO} &   \cellcolor[HTML]{c6dfed}91.25   &   \cellcolor[HTML]{c9e1ee}36.18   &   \cellcolor[HTML]{d2e6f1}3.50   &   \cellcolor[HTML]{e4f0f6}71.43   &   \cellcolor[HTML]{e7f1f7}27.97   &   \cellcolor[HTML]{e8f2f8}11.00   &   \cellcolor[HTML]{fc8c59}59.92   &   \cellcolor[HTML]{fcad89}67.21   &   \cellcolor[HTML]{fdbd9f}72.45  \\ 
 \textit{w/rand} &   \cellcolor[HTML]{c9e1ee}91.04   &   \cellcolor[HTML]{dcecf4}35.34   &   \cellcolor[HTML]{d2e6f0}3.48   &   \cellcolor[HTML]{fcfdfe}69.40   &   \cellcolor[HTML]{feeee6}26.22   &   \cellcolor[HTML]{d4e7f1}10.12   &   \cellcolor[HTML]{fc8c59}56.54   &   \cellcolor[HTML]{c3ddec}79.47   &   \cellcolor[HTML]{fdd0bb}74.11  \\ 
 \textit{w/atom} &   \cellcolor[HTML]{d2e6f1}90.22   &   \cellcolor[HTML]{e0edf5}35.20   &   \cellcolor[HTML]{d9eaf3}3.80   &   \cellcolor[HTML]{f9fcfd}69.58   &   \cellcolor[HTML]{fef2ed}26.42   &   \cellcolor[HTML]{d1e5f0}9.98   &   \cellcolor[HTML]{c8e0ed}92.41   &   \cellcolor[HTML]{f2f7fa}75.42   &   \cellcolor[HTML]{d4e7f1}\textbf{81.91}  \\ 
  \hline \multicolumn{1}{l|}{+RSO} &   \cellcolor[HTML]{d7e8f2}89.82   &   \cellcolor[HTML]{b4d5e7}37.09   &   \cellcolor[HTML]{e3eff6}4.25   &   \cellcolor[HTML]{eaf3f8}70.93   &   \cellcolor[HTML]{dbebf3}28.49   &   \cellcolor[HTML]{f4f9fb}11.53   &   \cellcolor[HTML]{fc8c59}47.26   &   \cellcolor[HTML]{fdd7c5}70.84   &   \cellcolor[HTML]{fc9d72}69.71  \\ 
 \textit{w/rand} &   \cellcolor[HTML]{d4e7f1}90.09   &   \cellcolor[HTML]{e1eef5}35.14   &   \cellcolor[HTML]{dfedf5}4.05   &   \cellcolor[HTML]{f8fbfd}69.67   &   \cellcolor[HTML]{fef4ef}26.49   &   \cellcolor[HTML]{daeaf3}10.39   &   \cellcolor[HTML]{fc8c59}37.45   &   \cellcolor[HTML]{bbd9e9}80.18   &   \cellcolor[HTML]{fc996c}69.35  \\ 
 \textit{w/atom} &   \cellcolor[HTML]{ebf3f8}88.11   &   \cellcolor[HTML]{fef9f7}33.64   &   \cellcolor[HTML]{ecf4f9}4.64   &   \cellcolor[HTML]{fcfdfe}69.40   &   \cellcolor[HTML]{feeee6}26.22   &   \cellcolor[HTML]{d4e7f1}10.12   &   \cellcolor[HTML]{fdbb9d}71.20   &   \cellcolor[HTML]{bad9e9}80.23   &   \cellcolor[HTML]{fef4ef}\textbf{77.23}

\\ \arrayrulecolor{black} 
\hline

\end{tabular}
}
\caption{Experimental results of models trained by different preference learning techniques. `FS' denotes FActScore, `NC' denotes the number of correct atomic facts and `NE' denotes the number of error atomic facts. `Acc' stands for `Accuracy' and `Rec' stands for 'Recall Score'. `Avg' is averaged across all `FS' and `Acc' values. Cells in blue indicate a performance increase, while cells in orange indicate a performance decrease.}
\label{our_method}
\end{table*}

\subsection{Experimental Results}
We create 2063 atomic preferences each for LLaMA-2-7B-Chat and LLaMA-3-8B-Instruct, and train the models using both these atomic preferences and the previously generated general preferences. The training process and the evaluation of the models follow the procedure outlined in Section \ref{general_preference_evaluation}. The experimental results are shown in Table \ref{our_method} (our method is denoted as `\textit{w/atom}') and we can draw the following observations: 

(1) Add atomic preferences to the original general preferences gain performance increase among almost all the learning algorithms. For example, for each preference learning algorithm, our method achieves improvements compared with the general-preference-only baselines. This demonstrates the effectiveness of the atomic preferences in contributing to the enhancement of facuality.


(2) Despite containing knowledge of only one single factoid, the atomic preferences can enhance the factuality in long-form responses. For example, the factuality of LLaMA-2 on the FAVA dataset improves after the incorporation of atomic preferences compared to using only the general preferences.

\subsection{Ablation Study}
To demonstrate the effectiveness of the selection of atomic facts in the general preferences, we conduct the following ablation study.  We randomly select questions from the TriviaQA dataset \citep{joshi-etal-2017-triviaqa}. To align with the atomic preferences, we filter these questions to ensure that they are relevant to the famous individuals in the training data and also contain a single `potentially known' fact according to stochastically sampled responses. Therefore, we use each question and its corresponding contradicted responses to construct preference pairs directly. We train the models based on these preferences and the previously generated general preferences (denoted as `\textit{w/random}') and the experimental results are shown in Table \ref{our_method}.


(1) Preferences constructed with the randomly selected questions are not as effective as those constructed with atomic facts selected from the general preferences. For example, the average accuracy of the `\textit{w/random}' method does not exceed that of our `\textit{w/atom}' method across almost all the preference learning algorithms and both models.

(2) Randomly selected QA samples are more beneficial for short-form factuality but contribute less to longer responses on the scale of several sentences or paragraphs. For example, while the accuracy on the KUQA dataset is generally slightly higher than that achieved with our method, it slightly drops on the other three datasets.

\section{Discussion}

In this section, we further explore whether merely increasing the quantity or improving the quality of training preference pairs facilitates the preference learning of factuality more effectively. Our main finding is that scaling or a greater difference in preference pairs does not necessarily result in improved factuality performance.





\begin{figure*}[]
\begin{center}
\resizebox{\linewidth}{!}{
\includegraphics[]{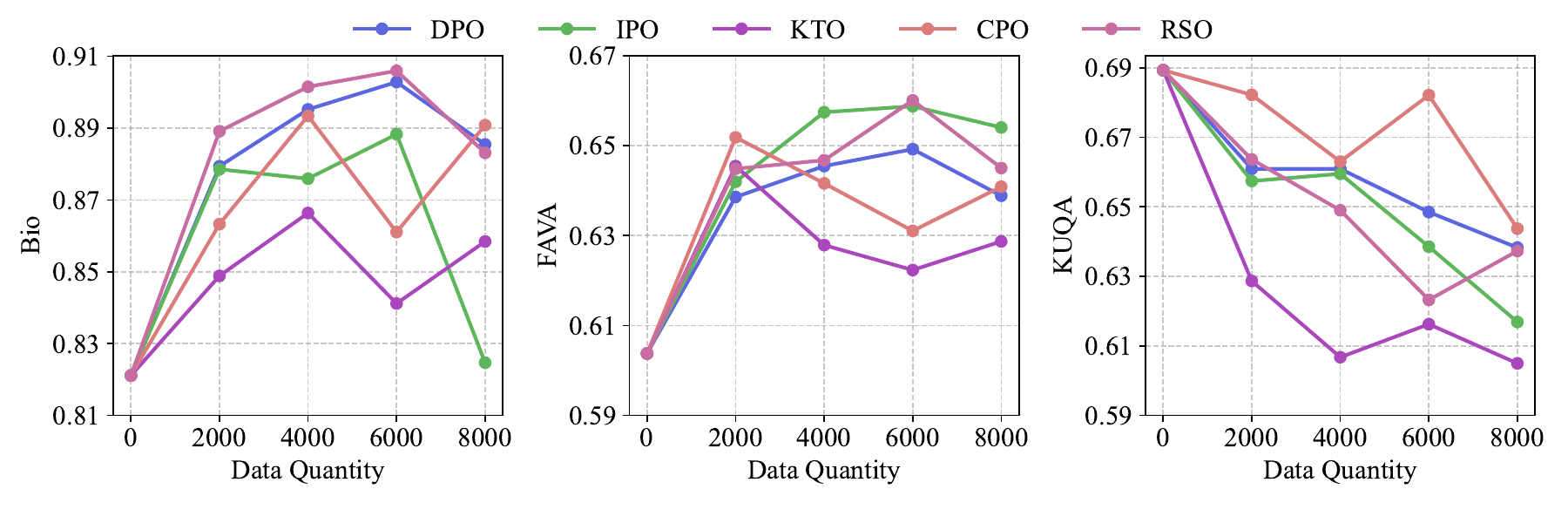} 
}
\caption{Experimental results of models trained by various number of training preferences.}
\label{quantity}
\end{center}
\end{figure*}

 \begin{figure}[]
\begin{center}
\resizebox{\linewidth}{!}{
\includegraphics[]{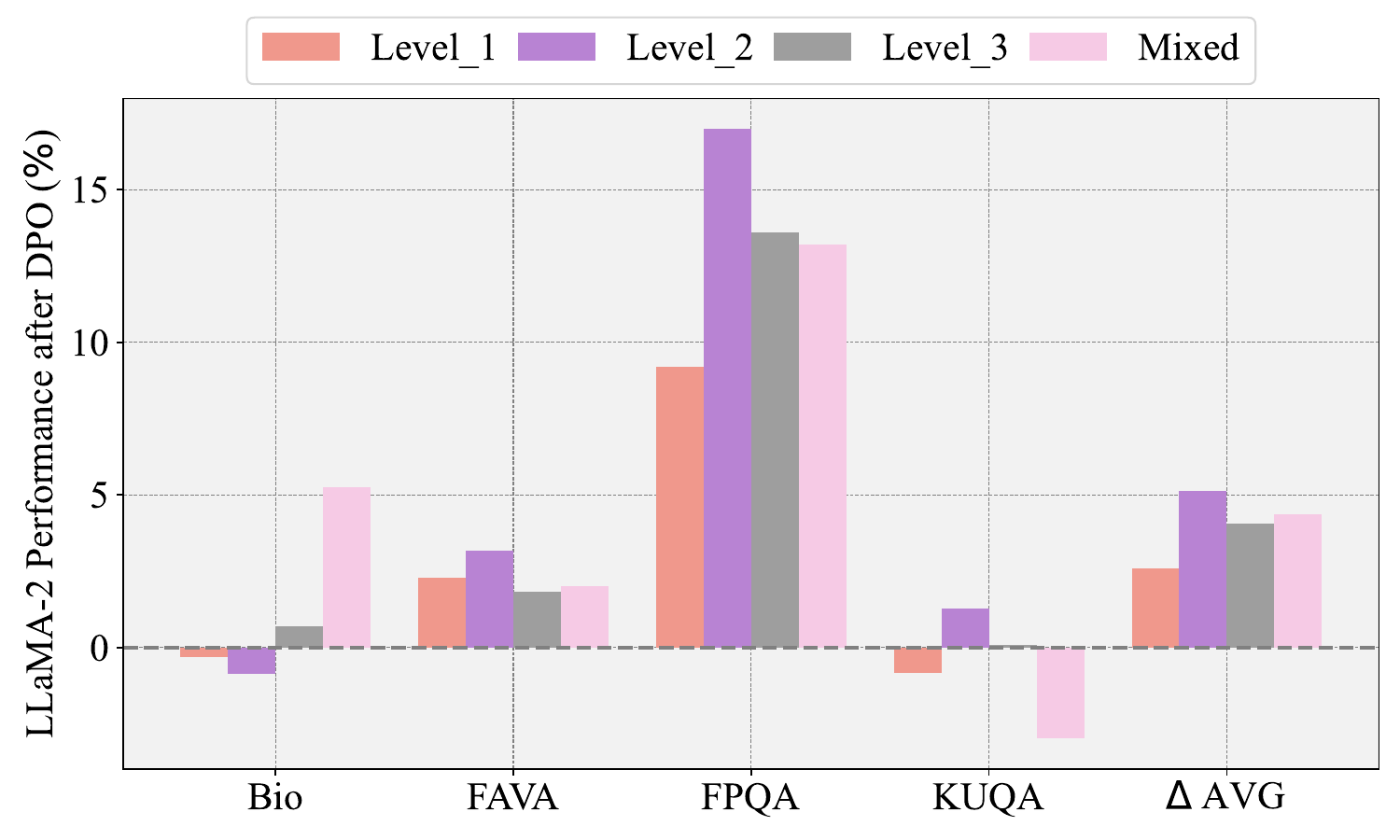} 
}
\caption{Experimental results of models trained by preferences of different quality levels using DPO. More results are presented in Appendix \ref{sec:data_quality_results}.}
\label{quality}
\end{center}
\end{figure}

\subsection{Data Quantity}
Initially, we investigate the influence of increasing the number of the training preference pairs. We create different amount of preference pairs and train LLaMA-2-7B-Chat based on them. The experimental results are shown in Figure \ref{quantity}.

We can draw the following observation: scaling is ineffective when training models for factuality. For example, what can be clearly seen in the figure is the dramatic decline in performance on the KUQA dataset as the number of training preference pairs increase. For the Bio and FAVA datasets, the performance quickly reaches a peak and then struggles to improve further. It suggests that a moderate amount of preference pairs is sufficient for the model to learn factuality, and simply adding more data might even backfire.

\subsection{Data Quality}
Subsequently, we study the influence of data quality in training preference pairs. We define the preference data quality as the difference in factuality scores between the preferred and dispreferred responses. Formally, the data quality score $q$ of a preference $(x,y_w,y_l)$ is calculated as $q=f(y_w) - f(y_l)$ where $f(y)$ represents the factuality score of model response $y$.

We divide the general preferences generated in Section \ref{general_preference_evaluation} into three levels of data groups according to their quality score: Level1 ($0<q\leq0.1$), Level2 ($1<q\leq0.2$) and Level3 ($q>0.2$). We have also added a mixed data group encompassing data of all quality levels.  To ensure a fair comparison, we ensure that each data group is filtered to have an equal quantity.  Then we use DPO to train LLaMA-2-7B-Chat using each data group. The experimental results are shown in Figure \ref{quality}.

We can draw the following observation: preference pairs of better quality do not necessarily ensure better performance. For example, data group with quality Level2 consistently outperforms others on OOD datasets. One possible reason is that the preference learning is actually aligning the knowledge inside the model parameters. Meanwhile, the quality of the preference pairs is assessed based on  whether the responses align with the external knowledge sources, which are not necessarily relevant to the model itself.

\section{Conclusion}
In this work, we firstly conduct a comprehensive evaluation of the factuality of different models on OOD queries tuned by various preference learning methods and demonstrate their ineffective performance. Subsequently, we conduct a token distribution analysis and reveal that the primary cause of the ineffectiveness is under-alignment rather than over-alignment. Finally, we propose APEFT, a framework that can enhance model's awareness of factuality at the granularity of individual facts. Extensive experiments demonstrate the effectiveness of our proposed APEFT.

\section*{Limitations}
While our work provides an in-depth analysis of how to enhance model's factuality on OOD queries, it is still subject to the following limitations:
\textbf{(1)} Our research primarily concentrates on how to enhance the factuality of LLMs through fine-tuning. However, it naturally raises concerns about whether the same process could adversely affect other capabilities that LLMs have already developed, such as mathematical reasoning, code completion and safety preservation. Future research could systematically investigate the changes in various capabilities of LLMs during the fine-tuning process.
\textbf{(2)} Our work does not delve into the internals of LLMs. It would be interesting to develop a more fine-grained understanding of behaviour changes within the LLMs induced by fine-tuning. For instance, is there a tiny region within LLMs focusing on the concept of factuality, and do the primary changes in parameters occur in this area? We leave the exploration for future work. 

\bibliography{custom}

\begin{thebibliography}{41}
\providecommand{\natexlab}[1]{#1}

\bibitem[{Azar et~al.(2024)Azar, Guo, Piot, Munos, Rowland, Valko, and Calandriello}]{DBLP:conf/aistats/AzarGPMRVC24}
Mohammad~Gheshlaghi Azar, Zhaohan~Daniel Guo, Bilal Piot, R{\'{e}}mi Munos, Mark Rowland, Michal Valko, and Daniele Calandriello. 2024.
\newblock \href {https://proceedings.mlr.press/v238/gheshlaghi-azar24a.html} {A general theoretical paradigm to understand learning from human preferences}.
\newblock In \emph{International Conference on Artificial Intelligence and Statistics, 2-4 May 2024, Palau de Congressos, Valencia, Spain}, volume 238 of \emph{Proceedings of Machine Learning Research}, pages 4447--4455. {PMLR}.

\bibitem[{Azar et~al.(2023)Azar, Rowland, Piot, Guo, Calandriello, Valko, and Munos}]{azar2023ipo}
Mohammad~Gheshlaghi Azar, Mark Rowland, Bilal Piot, Daniel Guo, Daniele Calandriello, Michal Valko, and Rémi Munos. 2023.
\newblock \href {https://arxiv.org/abs/2310.12036} {A general theoretical paradigm to understand learning from human preferences}.
\newblock \emph{Preprint}, arXiv:2310.12036.

\bibitem[{Azaria and Mitchell(2023)}]{azaria-mitchell-2023-internal}
Amos Azaria and Tom Mitchell. 2023.
\newblock \href {https://doi.org/10.18653/v1/2023.findings-emnlp.68} {The internal state of an {LLM} knows when it{'}s lying}.
\newblock In \emph{Findings of the Association for Computational Linguistics: EMNLP 2023}, pages 967--976, Singapore. Association for Computational Linguistics.

\bibitem[{Bai et~al.(2022)Bai, Jones, Ndousse, Askell, Chen, DasSarma, Drain, Fort, Ganguli, Henighan, Joseph, Kadavath, Kernion, Conerly, El-Showk, Elhage, Hatfield-Dodds, Hernandez, Hume, Johnston, Kravec, Lovitt, Nanda, Olsson, Amodei, Brown, Clark, McCandlish, Olah, Mann, and Kaplan}]{bai2022training}
Yuntao Bai, Andy Jones, Kamal Ndousse, Amanda Askell, Anna Chen, Nova DasSarma, Dawn Drain, Stanislav Fort, Deep Ganguli, Tom Henighan, Nicholas Joseph, Saurav Kadavath, Jackson Kernion, Tom Conerly, Sheer El-Showk, Nelson Elhage, Zac Hatfield-Dodds, Danny Hernandez, Tristan Hume, Scott Johnston, Shauna Kravec, Liane Lovitt, Neel Nanda, Catherine Olsson, Dario Amodei, Tom Brown, Jack Clark, Sam McCandlish, Chris Olah, Ben Mann, and Jared Kaplan. 2022.
\newblock \href {https://arxiv.org/abs/2204.05862} {Training a helpful and harmless assistant with reinforcement learning from human feedback}.
\newblock \emph{Preprint}, arXiv:2204.05862.

\bibitem[{Bubeck et~al.(2023)Bubeck, Chandrasekaran, Eldan, Gehrke, Horvitz, Kamar, Lee, Lee, Li, Lundberg, Nori, Palangi, Ribeiro, and Zhang}]{bubeck2023sparks_of_agi}
Sébastien Bubeck, Varun Chandrasekaran, Ronen Eldan, Johannes Gehrke, Eric Horvitz, Ece Kamar, Peter Lee, Yin~Tat Lee, Yuanzhi Li, Scott Lundberg, Harsha Nori, Hamid Palangi, Marco~Tulio Ribeiro, and Yi~Zhang. 2023.
\newblock \href {https://arxiv.org/abs/2303.12712} {Sparks of artificial general intelligence: Early experiments with gpt-4}.
\newblock \emph{Preprint}, arXiv:2303.12712.

\bibitem[{Burns et~al.(2023)Burns, Izmailov, Kirchner, Baker, Gao, Aschenbrenner, Chen, Ecoffet, Joglekar, Leike, Sutskever, and Wu}]{burns2023weaktostrong}
Collin Burns, Pavel Izmailov, Jan~Hendrik Kirchner, Bowen Baker, Leo Gao, Leopold Aschenbrenner, Yining Chen, Adrien Ecoffet, Manas Joglekar, Jan Leike, Ilya Sutskever, and Jeff Wu. 2023.
\newblock \href {https://arxiv.org/abs/2312.09390} {Weak-to-strong generalization: Eliciting strong capabilities with weak supervision}.
\newblock \emph{Preprint}, arXiv:2312.09390.

\bibitem[{Chen et~al.(2024{\natexlab{a}})Chen, Liu, Chen, Gu, Wu, Tao, Fu, and Ye}]{chen2024inside}
Chao Chen, Kai Liu, Ze~Chen, Yi~Gu, Yue Wu, Mingyuan Tao, Zhihang Fu, and Jieping Ye. 2024{\natexlab{a}}.
\newblock \href {https://arxiv.org/abs/2402.03744} {Inside: Llms' internal states retain the power of hallucination detection}.
\newblock \emph{Preprint}, arXiv:2402.03744.

\bibitem[{Chen et~al.(2024{\natexlab{b}})Chen, Jiao, Li, Qin, Ravaut, Zhao, Xiong, and Joty}]{chen2024chatgpts}
Hailin Chen, Fangkai Jiao, Xingxuan Li, Chengwei Qin, Mathieu Ravaut, Ruochen Zhao, Caiming Xiong, and Shafiq Joty. 2024{\natexlab{b}}.
\newblock \href {https://arxiv.org/abs/2311.16989} {Chatgpt's one-year anniversary: Are open-source large language models catching up?}
\newblock \emph{Preprint}, arXiv:2311.16989.

\bibitem[{Chen et~al.(2024{\natexlab{c}})Chen, Song, and Li}]{chen2024grath}
Weixin Chen, Dawn Song, and Bo~Li. 2024{\natexlab{c}}.
\newblock \href {https://arxiv.org/abs/2401.12292} {Grath: Gradual self-truthifying for large language models}.
\newblock \emph{Preprint}, arXiv:2401.12292.

\bibitem[{Chuang et~al.(2023)Chuang, Xie, Luo, Kim, Glass, and He}]{chuang2023dola}
Yung-Sung Chuang, Yujia Xie, Hongyin Luo, Yoon Kim, James Glass, and Pengcheng He. 2023.
\newblock Dola: Decoding by contrasting layers improves factuality in large language models.
\newblock \emph{arXiv preprint arXiv:2309.03883}.

\bibitem[{Cui et~al.(2023)Cui, Li, Yan, Chen, and Yuan}]{cui2023chatlaw}
Jiaxi Cui, Zongjian Li, Yang Yan, Bohua Chen, and Li~Yuan. 2023.
\newblock \href {https://arxiv.org/abs/2306.16092} {Chatlaw: Open-source legal large language model with integrated external knowledge bases}.
\newblock \emph{Preprint}, arXiv:2306.16092.

\bibitem[{Ethayarajh et~al.(2024)Ethayarajh, Xu, Muennighoff, Jurafsky, and Kiela}]{ethayarajh2024kto}
Kawin Ethayarajh, Winnie Xu, Niklas Muennighoff, Dan Jurafsky, and Douwe Kiela. 2024.
\newblock \href {https://arxiv.org/abs/2402.01306} {Kto: Model alignment as prospect theoretic optimization}.
\newblock \emph{Preprint}, arXiv:2402.01306.

\bibitem[{Gou et~al.(2023)Gou, Shao, Gong, Shen, Yang, Duan, and Chen}]{gou2023critic}
Zhibin Gou, Zhihong Shao, Yeyun Gong, Yelong Shen, Yujiu Yang, Nan Duan, and Weizhu Chen. 2023.
\newblock \href {https://arxiv.org/abs/2305.11738} {Critic: Large language models can self-correct with tool-interactive critiquing}.
\newblock \emph{Preprint}, arXiv:2305.11738.

\bibitem[{Hosking et~al.(2024)Hosking, Blunsom, and Bartolo}]{hosking2024human}
Tom Hosking, Phil Blunsom, and Max Bartolo. 2024.
\newblock \href {https://arxiv.org/abs/2309.16349} {Human feedback is not gold standard}.
\newblock \emph{Preprint}, arXiv:2309.16349.

\bibitem[{Huang et~al.(2023)Huang, Yu, Ma, Zhong, Feng, Wang, Chen, Peng, Feng, Qin, and Liu}]{huang2023survey_of_hallucination}
Lei Huang, Weijiang Yu, Weitao Ma, Weihong Zhong, Zhangyin Feng, Haotian Wang, Qianglong Chen, Weihua Peng, Xiaocheng Feng, Bing Qin, and Ting Liu. 2023.
\newblock \href {https://arxiv.org/abs/2311.05232} {A survey on hallucination in large language models: Principles, taxonomy, challenges, and open questions}.
\newblock \emph{Preprint}, arXiv:2311.05232.

\bibitem[{Jin et~al.(2024)Jin, Cao, Wang, He, Yuan, Li, Chen, Liu, and Zhao}]{jin2024rwku}
Zhuoran Jin, Pengfei Cao, Chenhao Wang, Zhitao He, Hongbang Yuan, Jiachun Li, Yubo Chen, Kang Liu, and Jun Zhao. 2024.
\newblock \href {https://arxiv.org/abs/2406.10890} {Rwku: Benchmarking real-world knowledge unlearning for large language models}.
\newblock \emph{Preprint}, arXiv:2406.10890.

\bibitem[{Joshi et~al.(2017)Joshi, Choi, Weld, and Zettlemoyer}]{joshi-etal-2017-triviaqa}
Mandar Joshi, Eunsol Choi, Daniel Weld, and Luke Zettlemoyer. 2017.
\newblock \href {https://doi.org/10.18653/v1/P17-1147} {{T}rivia{QA}: A large scale distantly supervised challenge dataset for reading comprehension}.
\newblock In \emph{Proceedings of the 55th Annual Meeting of the Association for Computational Linguistics (Volume 1: Long Papers)}, pages 1601--1611, Vancouver, Canada. Association for Computational Linguistics.

\bibitem[{Li et~al.(2023)Li, Patel, Vi{\'{e}}gas, Pfister, and Wattenberg}]{li2023inferencetime}
Kenneth Li, Oam Patel, Fernanda~B. Vi{\'{e}}gas, Hanspeter Pfister, and Martin Wattenberg. 2023.
\newblock \href {http://papers.nips.cc/paper\_files/paper/2023/hash/81b8390039b7302c909cb769f8b6cd93-Abstract-Conference.html} {Inference-time intervention: Eliciting truthful answers from a language model}.
\newblock In \emph{Advances in Neural Information Processing Systems 36: Annual Conference on Neural Information Processing Systems 2023, NeurIPS 2023, New Orleans, LA, USA, December 10 - 16, 2023}.

\bibitem[{Lin et~al.(2023)Lin, Ravichander, Lu, Dziri, Sclar, Chandu, Bhagavatula, and Choi}]{lin2023unlocking}
Bill~Yuchen Lin, Abhilasha Ravichander, Ximing Lu, Nouha Dziri, Melanie Sclar, Khyathi Chandu, Chandra Bhagavatula, and Yejin Choi. 2023.
\newblock \href {https://arxiv.org/abs/2312.01552} {The unlocking spell on base llms: Rethinking alignment via in-context learning}.
\newblock \emph{Preprint}, arXiv:2312.01552.

\bibitem[{Lin et~al.(2024)Lin, Gao, Oguz, Xiong, Lin, tau Yih, and Chen}]{lin2024flame}
Sheng-Chieh Lin, Luyu Gao, Barlas Oguz, Wenhan Xiong, Jimmy Lin, Wen tau Yih, and Xilun Chen. 2024.
\newblock \href {https://arxiv.org/abs/2405.01525} {Flame: Factuality-aware alignment for large language models}.
\newblock \emph{Preprint}, arXiv:2405.01525.

\bibitem[{Liu et~al.(2024{\natexlab{a}})Liu, Xue, Chen, Chen, Zhao, Wang, Hou, Li, and Peng}]{survey_of_hallucination_in_VLLMs}
Hanchao Liu, Wenyuan Xue, Yifei Chen, Dapeng Chen, Xiutian Zhao, Ke~Wang, Liping Hou, Rongjun Li, and Wei Peng. 2024{\natexlab{a}}.
\newblock \href {https://doi.org/10.48550/ARXIV.2402.00253} {A survey on hallucination in large vision-language models}.
\newblock \emph{CoRR}, abs/2402.00253.

\bibitem[{Liu et~al.(2024{\natexlab{b}})Liu, Zhao, Joshi, Khalman, Saleh, Liu, and Liu}]{liu2024statistical}
Tianqi Liu, Yao Zhao, Rishabh Joshi, Misha Khalman, Mohammad Saleh, Peter~J. Liu, and Jialu Liu. 2024{\natexlab{b}}.
\newblock \href {https://arxiv.org/abs/2309.06657} {Statistical rejection sampling improves preference optimization}.
\newblock \emph{Preprint}, arXiv:2309.06657.

\bibitem[{Liu et~al.(2024{\natexlab{c}})Liu, Zeng, He, Jiang, and He}]{liu2024makes}
Wei Liu, Weihao Zeng, Keqing He, Yong Jiang, and Junxian He. 2024{\natexlab{c}}.
\newblock \href {https://arxiv.org/abs/2312.15685} {What makes good data for alignment? a comprehensive study of automatic data selection in instruction tuning}.
\newblock \emph{Preprint}, arXiv:2312.15685.

\bibitem[{Manakul et~al.(2023)Manakul, Liusie, and Gales}]{manakul-etal-2023-selfcheckgpt}
Potsawee Manakul, Adian Liusie, and Mark Gales. 2023.
\newblock \href {https://doi.org/10.18653/v1/2023.emnlp-main.557} {{S}elf{C}heck{GPT}: Zero-resource black-box hallucination detection for generative large language models}.
\newblock In \emph{Proceedings of the 2023 Conference on Empirical Methods in Natural Language Processing}, pages 9004--9017, Singapore. Association for Computational Linguistics.

\bibitem[{Min et~al.(2023)Min, Krishna, Lyu, Lewis, Yih, Koh, Iyyer, Zettlemoyer, and Hajishirzi}]{min-etal-2023-factscore}
Sewon Min, Kalpesh Krishna, Xinxi Lyu, Mike Lewis, Wen-tau Yih, Pang Koh, Mohit Iyyer, Luke Zettlemoyer, and Hannaneh Hajishirzi. 2023.
\newblock \href {https://doi.org/10.18653/v1/2023.emnlp-main.741} {{FA}ct{S}core: Fine-grained atomic evaluation of factual precision in long form text generation}.
\newblock In \emph{Proceedings of the 2023 Conference on Empirical Methods in Natural Language Processing}, pages 12076--12100, Singapore. Association for Computational Linguistics.

\bibitem[{Mishra et~al.(2024)Mishra, Asai, Balachandran, Wang, Neubig, Tsvetkov, and Hajishirzi}]{mishra2024finegrained}
Abhika Mishra, Akari Asai, Vidhisha Balachandran, Yizhong Wang, Graham Neubig, Yulia Tsvetkov, and Hannaneh Hajishirzi. 2024.
\newblock \href {https://arxiv.org/abs/2401.06855} {Fine-grained hallucination detection and editing for language models}.
\newblock \emph{Preprint}, arXiv:2401.06855.

\bibitem[{Mündler et~al.(2023)Mündler, He, Jenko, and Vechev}]{mündler2023selfcontradictory}
Niels Mündler, Jingxuan He, Slobodan Jenko, and Martin Vechev. 2023.
\newblock \href {https://arxiv.org/abs/2305.15852} {Self-contradictory hallucinations of large language models: Evaluation, detection and mitigation}.
\newblock \emph{Preprint}, arXiv:2305.15852.

\bibitem[{Ouyang et~al.(2022)Ouyang, Wu, Jiang, Almeida, Wainwright, Mishkin, Zhang, Agarwal, Slama, Ray, Schulman, Hilton, Kelton, Miller, Simens, Askell, Welinder, Christiano, Leike, and Lowe}]{ouyang2022training}
Long Ouyang, Jeffrey Wu, Xu~Jiang, Diogo Almeida, Carroll~L. Wainwright, Pamela Mishkin, Chong Zhang, Sandhini Agarwal, Katarina Slama, Alex Ray, John Schulman, Jacob Hilton, Fraser Kelton, Luke Miller, Maddie Simens, Amanda Askell, Peter Welinder, Paul~F. Christiano, Jan Leike, and Ryan Lowe. 2022.
\newblock \href {http://papers.nips.cc/paper\_files/paper/2022/hash/b1efde53be364a73914f58805a001731-Abstract-Conference.html} {Training language models to follow instructions with human feedback}.
\newblock In \emph{Advances in Neural Information Processing Systems 35: Annual Conference on Neural Information Processing Systems 2022, NeurIPS 2022, New Orleans, LA, USA, November 28 - December 9, 2022}.

\bibitem[{Rafailov et~al.(2023)Rafailov, Sharma, Mitchell, Manning, Ermon, and Finn}]{DBLP:conf/nips/RafailovSMMEF23}
Rafael Rafailov, Archit Sharma, Eric Mitchell, Christopher~D. Manning, Stefano Ermon, and Chelsea Finn. 2023.
\newblock \href {http://papers.nips.cc/paper\_files/paper/2023/hash/a85b405ed65c6477a4fe8302b5e06ce7-Abstract-Conference.html} {Direct preference optimization: Your language model is secretly a reward model}.
\newblock In \emph{Advances in Neural Information Processing Systems 36: Annual Conference on Neural Information Processing Systems 2023, NeurIPS 2023, New Orleans, LA, USA, December 10 - 16, 2023}.

\bibitem[{Sanh et~al.(2022)Sanh, Webson, Raffel, Bach, Sutawika, Alyafeai, Chaffin, Stiegler, Raja, Dey, Bari, Xu, Thakker, Sharma, Szczechla, Kim, Chhablani, Nayak, Datta, Chang, Jiang, Wang, Manica, Shen, Yong, Pandey, Bawden, Wang, Neeraj, Rozen, Sharma, Santilli, F{\'{e}}vry, Fries, Teehan, Scao, Biderman, Gao, Wolf, and Rush}]{DBLP:conf/iclr/SanhWRBSACSRDBX22}
Victor Sanh, Albert Webson, Colin Raffel, Stephen~H. Bach, Lintang Sutawika, Zaid Alyafeai, Antoine Chaffin, Arnaud Stiegler, Arun Raja, Manan Dey, M~Saiful Bari, Canwen Xu, Urmish Thakker, Shanya~Sharma Sharma, Eliza Szczechla, Taewoon Kim, Gunjan Chhablani, Nihal~V. Nayak, Debajyoti Datta, Jonathan Chang, Mike~Tian{-}Jian Jiang, Han Wang, Matteo Manica, Sheng Shen, Zheng~Xin Yong, Harshit Pandey, Rachel Bawden, Thomas Wang, Trishala Neeraj, Jos Rozen, Abheesht Sharma, Andrea Santilli, Thibault F{\'{e}}vry, Jason~Alan Fries, Ryan Teehan, Teven~Le Scao, Stella Biderman, Leo Gao, Thomas Wolf, and Alexander~M. Rush. 2022.
\newblock \href {https://openreview.net/forum?id=9Vrb9D0WI4} {Multitask prompted training enables zero-shot task generalization}.
\newblock In \emph{The Tenth International Conference on Learning Representations, {ICLR} 2022, Virtual Event, April 25-29, 2022}. OpenReview.net.

\bibitem[{Saunders et~al.(2022)Saunders, Yeh, Wu, Bills, Ouyang, Ward, and Leike}]{saunders2022selfcritiquing}
William Saunders, Catherine Yeh, Jeff Wu, Steven Bills, Long Ouyang, Jonathan Ward, and Jan Leike. 2022.
\newblock \href {https://arxiv.org/abs/2206.05802} {Self-critiquing models for assisting human evaluators}.
\newblock \emph{Preprint}, arXiv:2206.05802.

\bibitem[{Tian et~al.(2023)Tian, Mitchell, Yao, Manning, and Finn}]{tian2023finetuning}
Katherine Tian, Eric Mitchell, Huaxiu Yao, Christopher~D. Manning, and Chelsea Finn. 2023.
\newblock \href {https://arxiv.org/abs/2311.08401} {Fine-tuning language models for factuality}.
\newblock \emph{Preprint}, arXiv:2311.08401.

\bibitem[{Touvron et~al.(2023)Touvron, Martin, Stone, Albert, Almahairi, Babaei, Bashlykov, Batra, Bhargava, Bhosale, Bikel, Blecher, Ferrer, Chen, Cucurull, Esiobu, Fernandes, Fu, Fu, Fuller, Gao, Goswami, Goyal, Hartshorn, Hosseini, Hou, Inan, Kardas, Kerkez, Khabsa, Kloumann, Korenev, Koura, Lachaux, Lavril, Lee, Liskovich, Lu, Mao, Martinet, Mihaylov, Mishra, Molybog, Nie, Poulton, Reizenstein, Rungta, Saladi, Schelten, Silva, Smith, Subramanian, Tan, Tang, Taylor, Williams, Kuan, Xu, Yan, Zarov, Zhang, Fan, Kambadur, Narang, Rodriguez, Stojnic, Edunov, and Scialom}]{touvron2023llama}
Hugo Touvron, Louis Martin, Kevin Stone, Peter Albert, Amjad Almahairi, Yasmine Babaei, Nikolay Bashlykov, Soumya Batra, Prajjwal Bhargava, Shruti Bhosale, Dan Bikel, Lukas Blecher, Cristian~Canton Ferrer, Moya Chen, Guillem Cucurull, David Esiobu, Jude Fernandes, Jeremy Fu, Wenyin Fu, Brian Fuller, Cynthia Gao, Vedanuj Goswami, Naman Goyal, Anthony Hartshorn, Saghar Hosseini, Rui Hou, Hakan Inan, Marcin Kardas, Viktor Kerkez, Madian Khabsa, Isabel Kloumann, Artem Korenev, Punit~Singh Koura, Marie-Anne Lachaux, Thibaut Lavril, Jenya Lee, Diana Liskovich, Yinghai Lu, Yuning Mao, Xavier Martinet, Todor Mihaylov, Pushkar Mishra, Igor Molybog, Yixin Nie, Andrew Poulton, Jeremy Reizenstein, Rashi Rungta, Kalyan Saladi, Alan Schelten, Ruan Silva, Eric~Michael Smith, Ranjan Subramanian, Xiaoqing~Ellen Tan, Binh Tang, Ross Taylor, Adina Williams, Jian~Xiang Kuan, Puxin Xu, Zheng Yan, Iliyan Zarov, Yuchen Zhang, Angela Fan, Melanie Kambadur, Sharan Narang, Aurelien Rodriguez, Robert Stojnic, Sergey Edunov, and Thomas
  Scialom. 2023.
\newblock \href {https://arxiv.org/abs/2307.09288} {Llama 2: Open foundation and fine-tuned chat models}.
\newblock \emph{Preprint}, arXiv:2307.09288.

\bibitem[{Trivedi et~al.(2023)Trivedi, Balasubramanian, Khot, and Sabharwal}]{trivedi-etal-2023-interleaving-retrieval}
Harsh Trivedi, Niranjan Balasubramanian, Tushar Khot, and Ashish Sabharwal. 2023.
\newblock \href {https://doi.org/10.18653/v1/2023.acl-long.557} {Interleaving retrieval with chain-of-thought reasoning for knowledge-intensive multi-step questions}.
\newblock In \emph{Proceedings of the 61st Annual Meeting of the Association for Computational Linguistics (Volume 1: Long Papers)}, pages 10014--10037, Toronto, Canada. Association for Computational Linguistics.

\bibitem[{Wei et~al.(2024)Wei, Yang, Song, Lu, Hu, Huang, Tran, Peng, Liu, Huang, Du, and Le}]{wei2024longform}
Jerry Wei, Chengrun Yang, Xinying Song, Yifeng Lu, Nathan Hu, Jie Huang, Dustin Tran, Daiyi Peng, Ruibo Liu, Da~Huang, Cosmo Du, and Quoc~V. Le. 2024.
\newblock \href {https://arxiv.org/abs/2403.18802} {Long-form factuality in large language models}.
\newblock \emph{Preprint}, arXiv:2403.18802.

\bibitem[{Xu et~al.(2024{\natexlab{a}})Xu, Sharaf, Chen, Tan, Shen, Durme, Murray, and Kim}]{xu2024cpo}
Haoran Xu, Amr Sharaf, Yunmo Chen, Weiting Tan, Lingfeng Shen, Benjamin~Van Durme, Kenton Murray, and Young~Jin Kim. 2024{\natexlab{a}}.
\newblock \href {https://arxiv.org/abs/2401.08417} {Contrastive preference optimization: Pushing the boundaries of llm performance in machine translation}.
\newblock \emph{Preprint}, arXiv:2401.08417.

\bibitem[{Xu et~al.(2024{\natexlab{b}})Xu, Sharaf, Chen, Tan, Shen, Durme, Murray, and Kim}]{xu2024contrastive}
Haoran Xu, Amr Sharaf, Yunmo Chen, Weiting Tan, Lingfeng Shen, Benjamin~Van Durme, Kenton Murray, and Young~Jin Kim. 2024{\natexlab{b}}.
\newblock \href {https://arxiv.org/abs/2401.08417} {Contrastive preference optimization: Pushing the boundaries of llm performance in machine translation}.
\newblock \emph{Preprint}, arXiv:2401.08417.

\bibitem[{Yuan et~al.(2024)Yuan, Cao, Jin, Chen, Zeng, Liu, and Zhao}]{yuan2024whispers}
Hongbang Yuan, Pengfei Cao, Zhuoran Jin, Yubo Chen, Daojian Zeng, Kang Liu, and Jun Zhao. 2024.
\newblock \href {https://arxiv.org/abs/2402.19103} {Whispers that shake foundations: Analyzing and mitigating false premise hallucinations in large language models}.
\newblock \emph{Preprint}, arXiv:2402.19103.

\bibitem[{Zhang et~al.(2023)Zhang, Li, Cui, Cai, Liu, Fu, Huang, Zhao, Zhang, Chen, Wang, Luu, Bi, Shi, and Shi}]{zhang2023sirens_song}
Yue Zhang, Yafu Li, Leyang Cui, Deng Cai, Lemao Liu, Tingchen Fu, Xinting Huang, Enbo Zhao, Yu~Zhang, Yulong Chen, Longyue Wang, Anh~Tuan Luu, Wei Bi, Freda Shi, and Shuming Shi. 2023.
\newblock \href {https://arxiv.org/abs/2309.01219} {Siren's song in the ai ocean: A survey on hallucination in large language models}.
\newblock \emph{Preprint}, arXiv:2309.01219.

\bibitem[{Zhou et~al.(2023)Zhou, Liu, Xu, Iyer, Sun, Mao, Ma, Efrat, Yu, Yu, Zhang, Ghosh, Lewis, Zettlemoyer, and Levy}]{less_is_more}
Chunting Zhou, Pengfei Liu, Puxin Xu, Srinivasan Iyer, Jiao Sun, Yuning Mao, Xuezhe Ma, Avia Efrat, Ping Yu, Lili Yu, Susan Zhang, Gargi Ghosh, Mike Lewis, Luke Zettlemoyer, and Omer Levy. 2023.
\newblock \href {http://papers.nips.cc/paper\_files/paper/2023/hash/ac662d74829e4407ce1d126477f4a03a-Abstract-Conference.html} {{LIMA:} less is more for alignment}.
\newblock In \emph{Advances in Neural Information Processing Systems 36: Annual Conference on Neural Information Processing Systems 2023, NeurIPS 2023, New Orleans, LA, USA, December 10 - 16, 2023}.

\bibitem[{Zou et~al.(2023)Zou, Phan, Chen, Campbell, Guo, Ren, Pan, Yin, Mazeika, Dombrowski, Goel, Li, Byun, Wang, Mallen, Basart, Koyejo, Song, Fredrikson, Kolter, and Hendrycks}]{zou2023transparency}
Andy Zou, Long Phan, Sarah Chen, James Campbell, Phillip Guo, Richard Ren, Alexander Pan, Xuwang Yin, Mantas Mazeika, Ann-Kathrin Dombrowski, Shashwat Goel, Nathaniel Li, Michael~J. Byun, Zifan Wang, Alex Mallen, Steven Basart, Sanmi Koyejo, Dawn Song, Matt Fredrikson, Zico Kolter, and Dan Hendrycks. 2023.
\newblock \href {https://arxiv.org/abs/2310.01405} {Representation engineering: A top-down approach to ai transparency}.
\newblock \emph{Preprint}, arXiv:2310.01405.

\end{thebibliography}

\appendix

\section{Preference Learning Algorithms}
\label{sec:preference_learning_algorithms}
In this section, we formally introduce the loss functions used by the various preference learning algorithms. Formally, given the preference dataset $D=\{x^i,y_w^i,y_l^i\}_{i=1}^{N}$, the reference policy model $\pi_\theta$ and the current policy model  $\pi_{\mathrm{ref}}$,  the loss functions can be articulated  as follows:

(1) \textbf{DPO} (\textbf{D}irect \textbf{P}reference \textbf{O}ptimization)
\begin{tiny}
\begin{equation*}
\begin{split}
    \mathcal{L}_{\mathrm{DPO}}=-\underset{\left(y_w, y_l, x\right) \sim D}{\mathbb{E}} \left[ \log \sigma\left(  \beta \log \frac{\pi_\theta\left(y_w \mid x\right)}{\pi_{\mathrm{ref}}\left(y_w \mid x\right)}  -\beta \log \frac{\pi_\theta\left(y_l \mid x\right)}{\pi_{\mathrm{ref}}\left(y_l \mid x\right)}\right) \right]
\end{split}
\end{equation*}
\end{tiny}

where $\beta$ is a hyperparameter controlling the deviation between the policy model  $\pi_\theta$ and the reference model  $\pi_{\mathrm{ref}}$during the optimization process. 

(2) \textbf{IPO} (\textbf{I}dentity \textbf{P}reference \textbf{O}ptimization)

\begin{small}
    \begin{equation*}
\mathcal{L}_{\mathrm{IPO}} =  \underset{\left(y_w, y_l, x\right) \sim D}{\mathbb{E}}\left(h_\pi\left(y_w, y_l, x\right)-\frac{\tau^{-1}}{2}\right)^2
\end{equation*}
\end{small}

\begin{small}
    \begin{equation*}
h_\pi\left(y, y^{\prime}, x\right)=\log \left(\frac{\pi(y \mid x) \pi_{\mathrm{ref}}\left(y^{\prime} \mid x\right)}{\pi\left(y^{\prime} \mid x\right) \pi_{\mathrm{ref}}(y \mid x)}\right)
\end{equation*}
\end{small}
 where $\tau$ is a regularisation term. A lower value of $\tau$ means a higher value of the log-likelihood ratio of $y_w$ to $y_l$.

(3) \textbf{KTO} (\textbf{K}ahneman-\textbf{T}versky Optimization)

\begin{small}
    \begin{equation*}
L_{\mathrm{KTO}}=\mathbb{E}_{x, y \sim D}\left[w(y)\left(1-v_{\mathrm{KTO}}(x, y ; \beta)\right)\right]
\end{equation*}
\end{small}

where 
\begin{small}
    \begin{equation*}
\begin{aligned}
r_{\mathrm{KTO}}(x, y) & =\beta \log \frac{\pi_\theta(y \mid x)}{\pi_{\mathrm{ref}}(y \mid x)} \\
z_{\mathrm{ref}} & =\mathbb{E}_{x^{\prime} \sim D}\left[\beta \operatorname{KL}\left(\pi_\theta\left(y^{\prime} \mid x^{\prime}\right) \| \pi_{\text {ref }}\left(y^{\prime} \mid x^{\prime}\right)\right)\right] \\
v_{\mathrm{KTO}}(x, y ; \beta) & =\left\{\begin{array}{l}
\sigma\left(r_{\mathrm{KTO}}(x, y)-z_{\text {ref }}\right) \text { if } y \sim y_{\text {desirable }} \mid x \\
\sigma\left(z_{\text {ref }}-r_{\mathrm{KTO}}(x, y)\right) \text { if } y \sim y_{\text {undesirable }} \mid x
\end{array}\right. \\
w(y) & = \begin{cases}\lambda_D & \text { if } y \sim y_{\text {desirable }} \mid x \\
\lambda_U & \text { if } y \sim y_{\text {undesirable }} \mid x\end{cases}
\end{aligned}
\end{equation*}
\end{small}

Notably, KTO only requires one response and a binary signal indicating whether it is `desirable' or `undesirable'. We set the \texttt{loss\_type} parameter to \texttt{kto\_pair} in \texttt{DPOTraine}r which is a highly simplified version and we leave a more sophisticated implementation for future work.

(4) \textbf{CPO} (\textbf{C}ontrastive \textbf{P}reference \textbf{O}ptimization)

\begin{small}
    \begin{equation*}
\begin{gathered}
\mathcal{L}_{\mathrm{NLL}}=-\mathbb{E}_{\left(x, y_w\right) \sim \mathcal{D}}\left[\log \pi_\theta\left(y_w \mid x\right)\right] \\
\mathcal{L}_{\text {prefer }}=-\mathbb{E}_{\left(x, y_w, y_l\right) \sim \mathcal{D}}\left[\operatorname { l o g } \sigma \left(\beta \log \pi_\theta\left(y_w \mid x\right)\right.\right. \\
\left.\left.\left.-\beta \log \pi_\theta\left(y_l \mid x\right)\right)\right)\right] \\
\mathcal{L}_{\mathrm{CPO}}=\mathcal{L}_{\text {prefer }}+\mathcal{L}_{\mathrm{NLL}}
\end{gathered}
\end{equation*}
\end{small}

where $\beta$ is a regularisation term. Notably, CPO doesn't require the reference model. Thus it is more memory-efficient and speed efficient. 

(5) \textbf{RSO} (\textbf{S}tatistical \textbf{R}ejection Sampling \textbf{O}ptimization)

\begin{tiny}
    \begin{equation*}
    \begin{split}
        \mathcal{L}_{\text{rso}}=\underset{\left(y_w, y_l, x\right) \sim D} {\mathbb{E}} \left[\max \left(0,1-\left[\gamma \log \frac{\pi_\theta\left(y_w \mid x\right)}{\pi_{\text{sft}}\left(y_w \mid x\right)}-\gamma \log \frac{\pi_\theta\left(y_l \mid x\right)}{\pi_{\text{sft}}\left(y_l \mid x\right)}\right]\right)\right]
    \end{split}
\end{equation*}
\end{tiny}
where $\gamma$ is a regularisation term.
In our work, the preferences are sampled and annotated directly from the models. Thus we use a simplified version by setting the supervised finetuning model to the original base reference policy model. This equals to substituting the sigmoid function in DPO loss with a hinge loss. We leave the original implementation for future work.

\section{Additional Results of General Preference Performance}
\label{sec:llama2_general_performance}
We present the performance change of LLaMA-2-7B-Chat trained by general preferences using various preference learning techniques.

\begin{figure}[t]
\centering
\includegraphics[width=\linewidth]{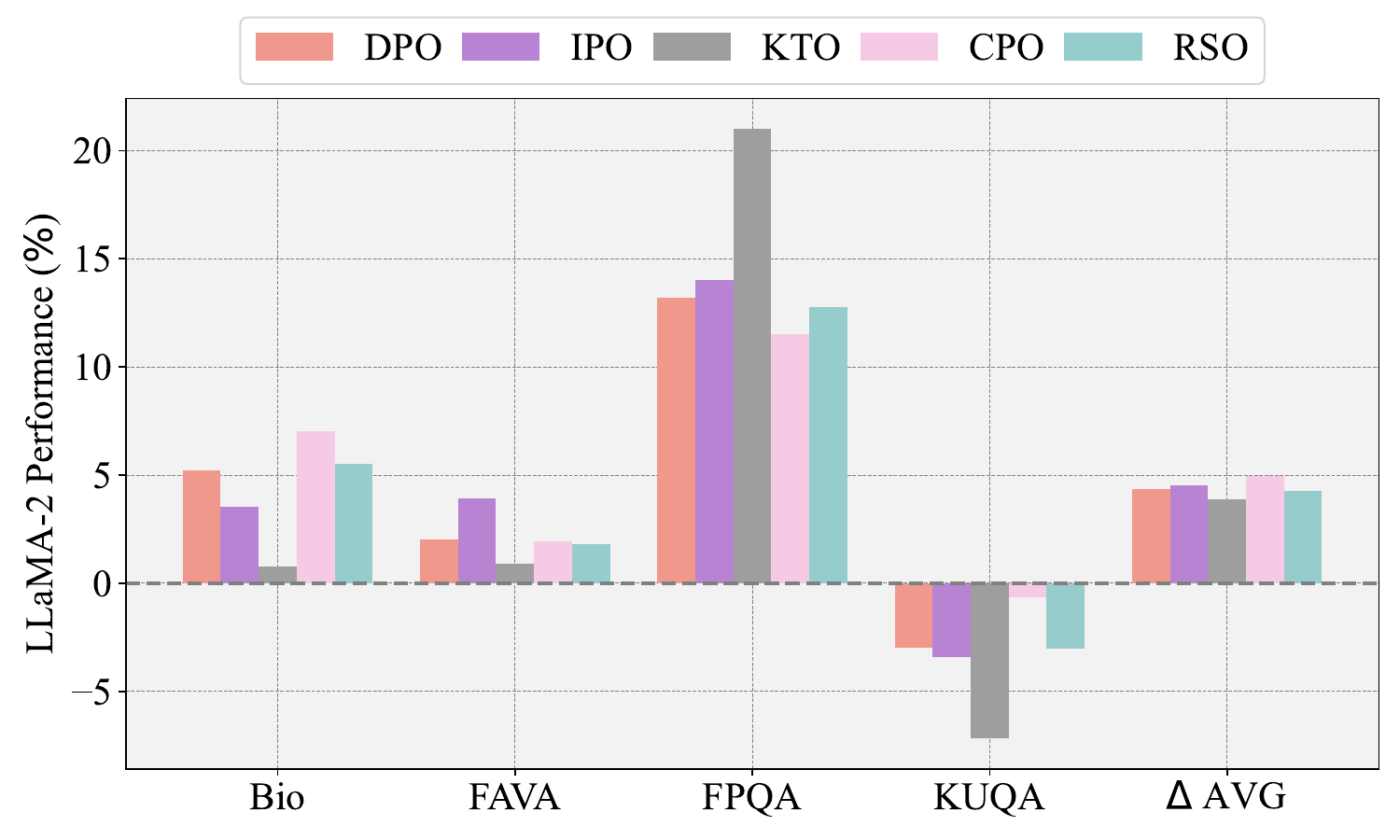} 
\caption{LLaMA-2-7B-Chat performance change before and after tuning using various preference learning techniques.}
\end{figure}

\section{Additional Result of Token Shift Analysis}
\label{sec:token_shift_analysis}
\begin{figure}[t]
\centering
\subfloat[Bio Dataset]{\includegraphics[width=\linewidth]{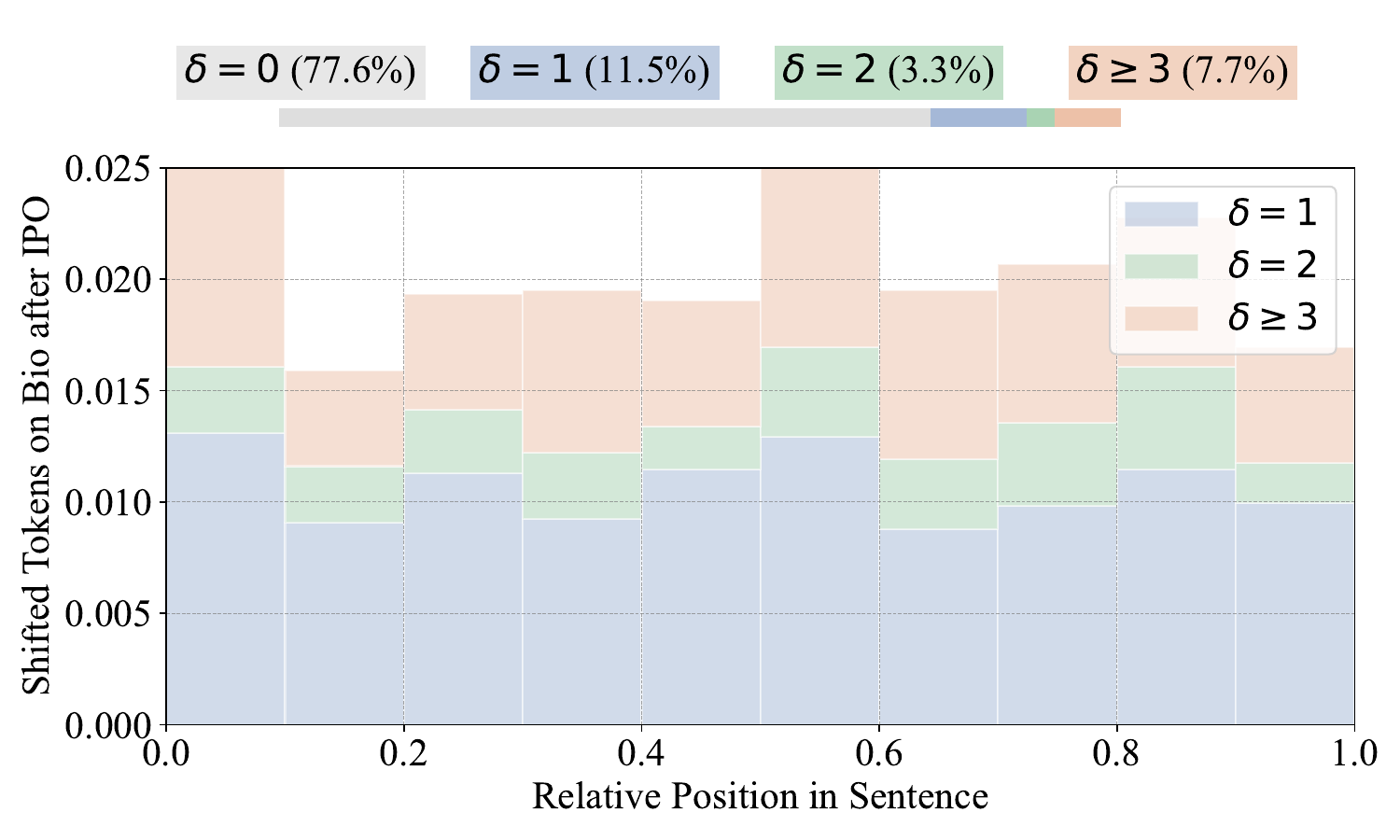}\label{attn_head_influence}} \hfill
\subfloat[FAVA Dataset]{\includegraphics[width=\linewidth]{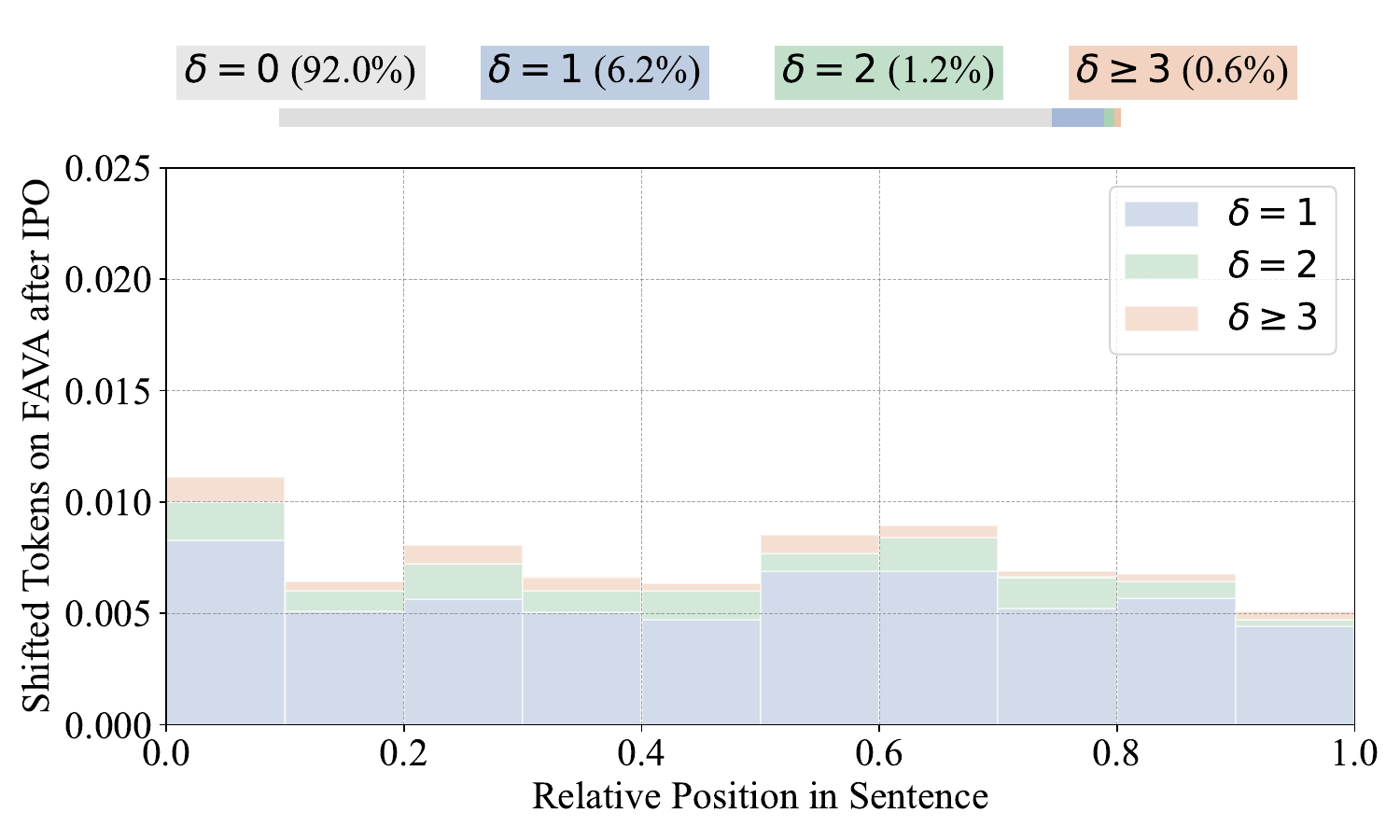}\label{attn_head_influence}} 
\caption{Token shift analysis on LLaMA-2 trained by IPO.}
\label{token_shift_result_ipo}
\end{figure}

\begin{figure}[t]
\centering
\subfloat[Bio Dataset]{\includegraphics[width=\linewidth]{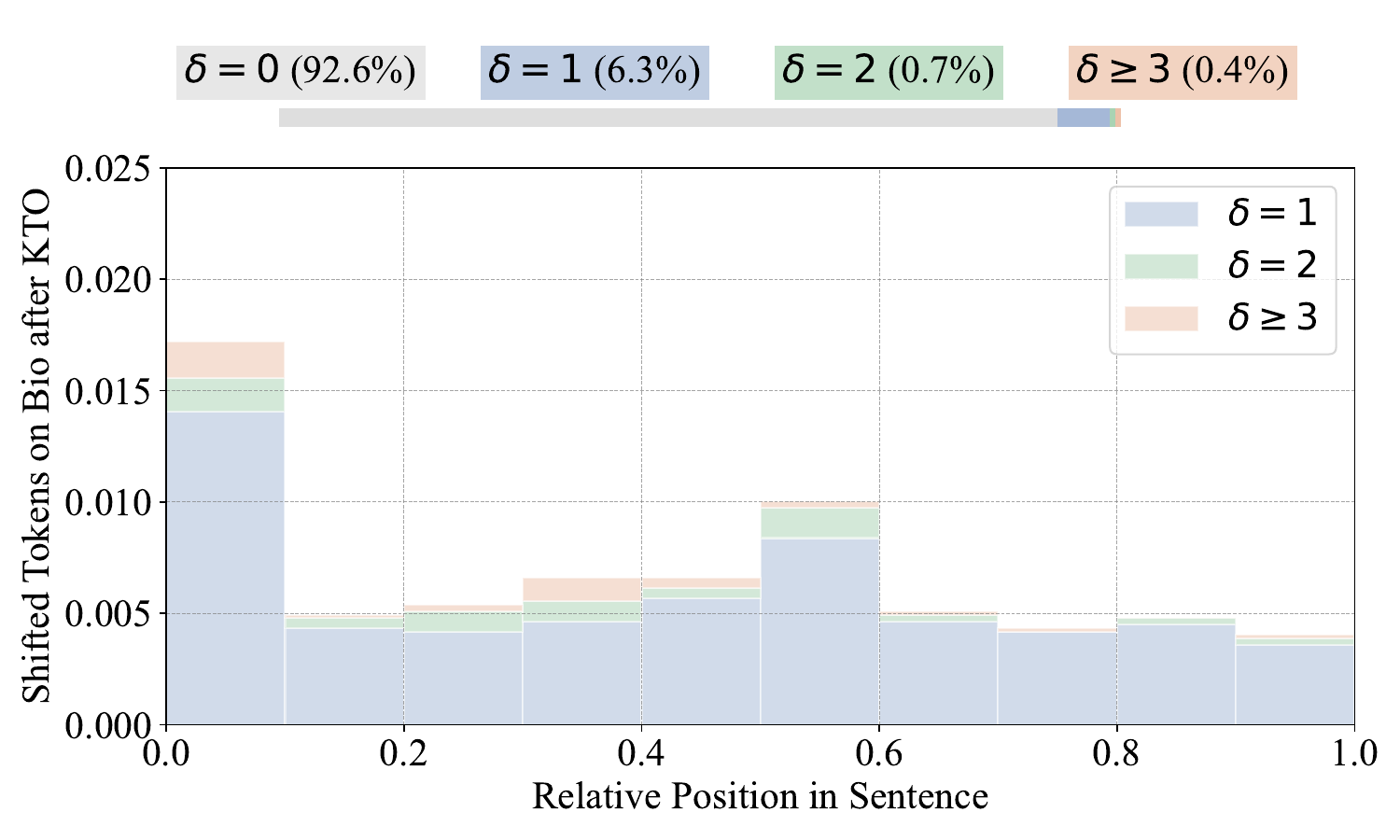}\label{attn_head_influence}} \hfill
\subfloat[FAVA Dataset]{\includegraphics[width=\linewidth]{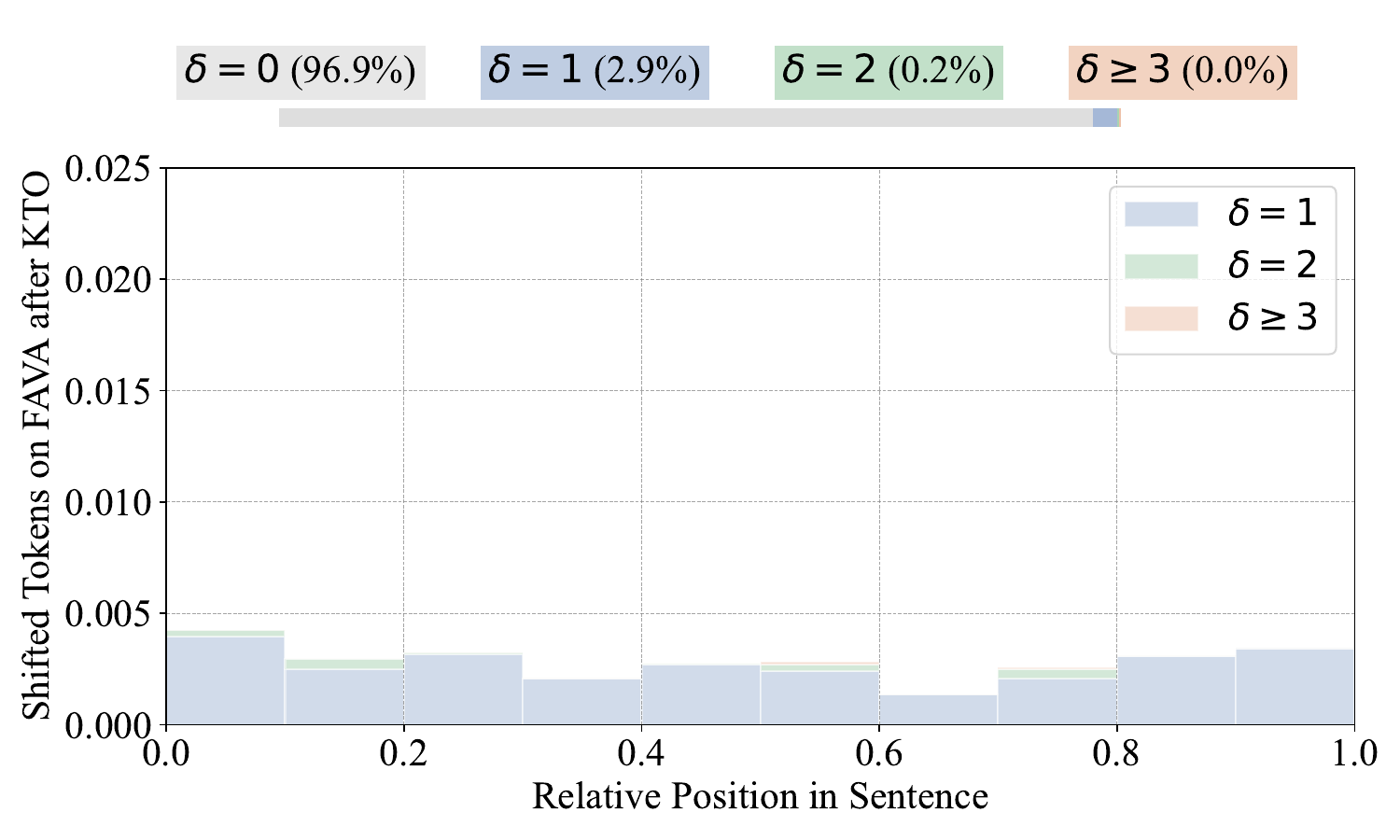}\label{attn_head_influence}} 
\caption{Token shift analysis on LLaMA-2 trained by KTO.}
\label{token_shift_result_kto}
\end{figure}

\begin{figure}[t]
\centering
\subfloat[Bio Dataset]{\includegraphics[width=\linewidth]{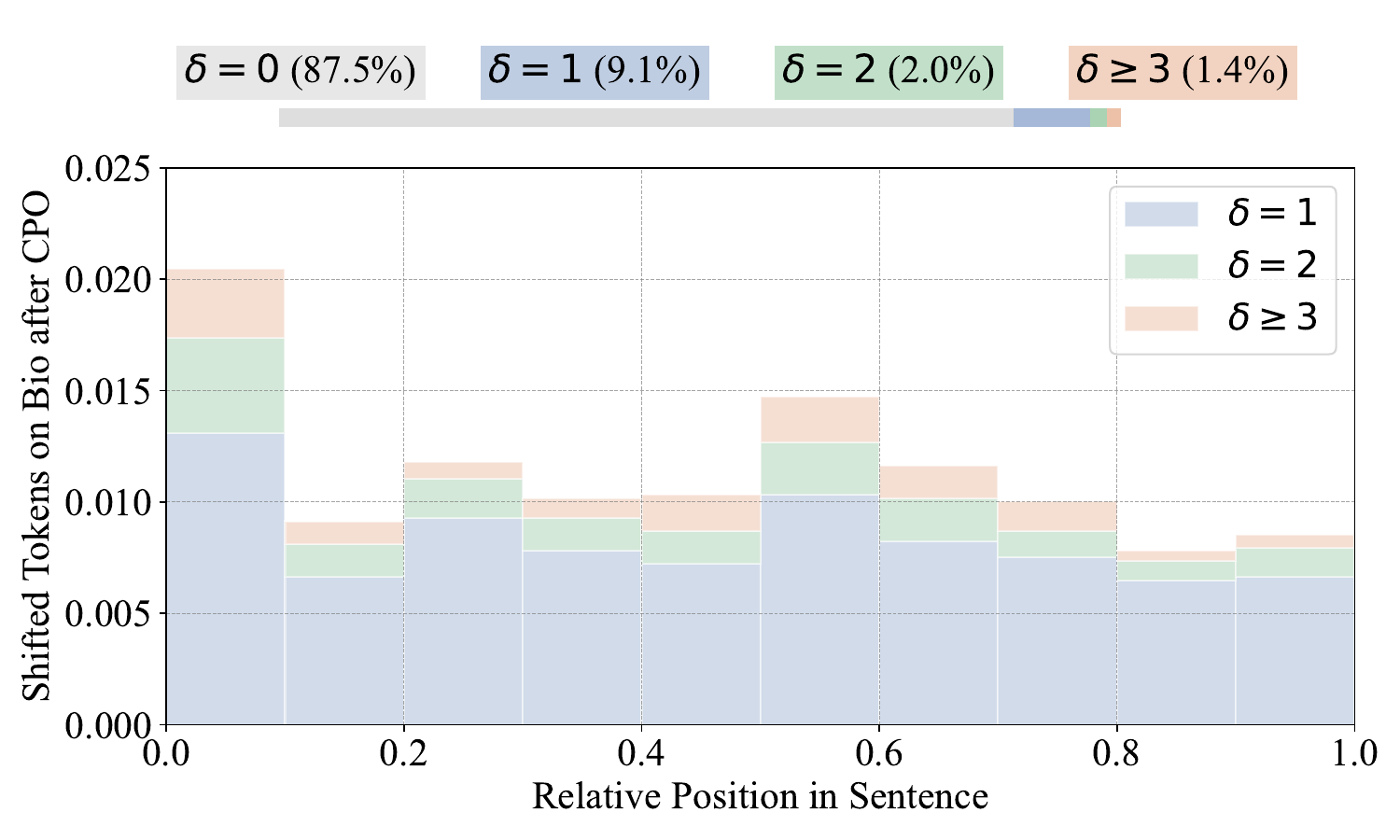}\label{attn_head_influence}} \hfill
\subfloat[FAVA Dataset]{\includegraphics[width=\linewidth]{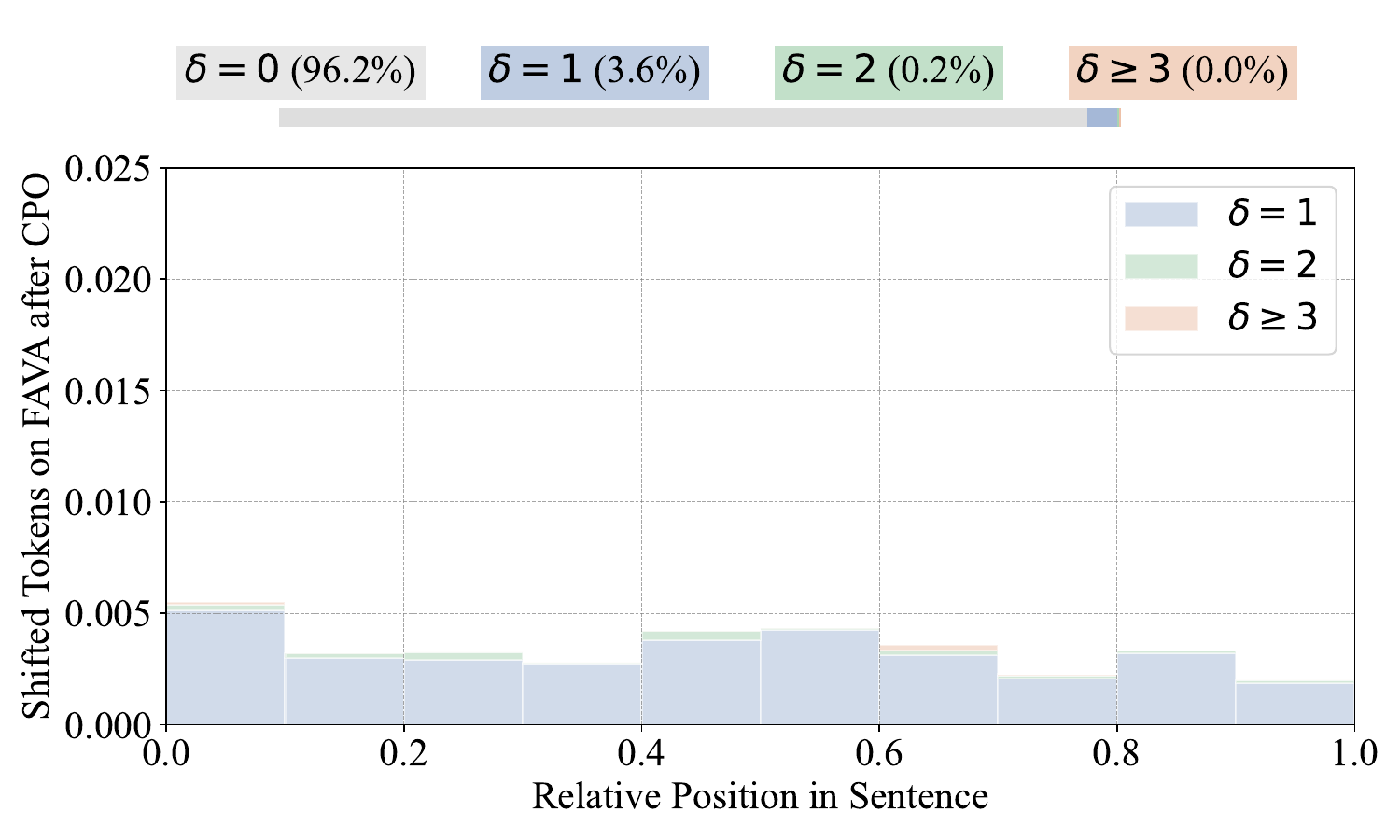}\label{attn_head_influence}} 
\caption{Token shift analysis on LLaMA-2 trained by CPO.}
\label{token_shift_result_cpo}
\end{figure}

\begin{figure}[t]
\centering
\subfloat[Bio Dataset]{\includegraphics[width=\linewidth]{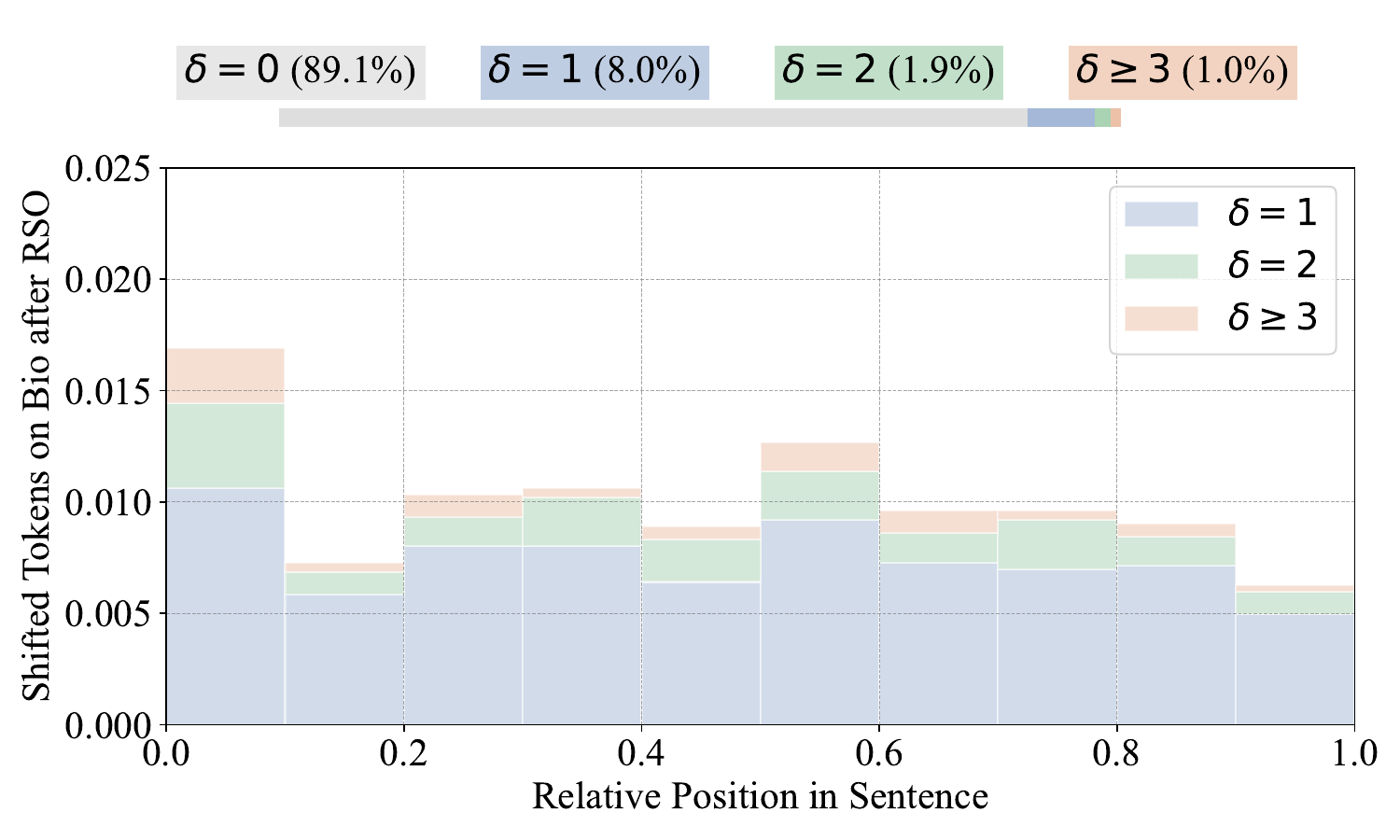}\label{attn_head_influence}} \hfill
\subfloat[FAVA Dataset]{\includegraphics[width=\linewidth]{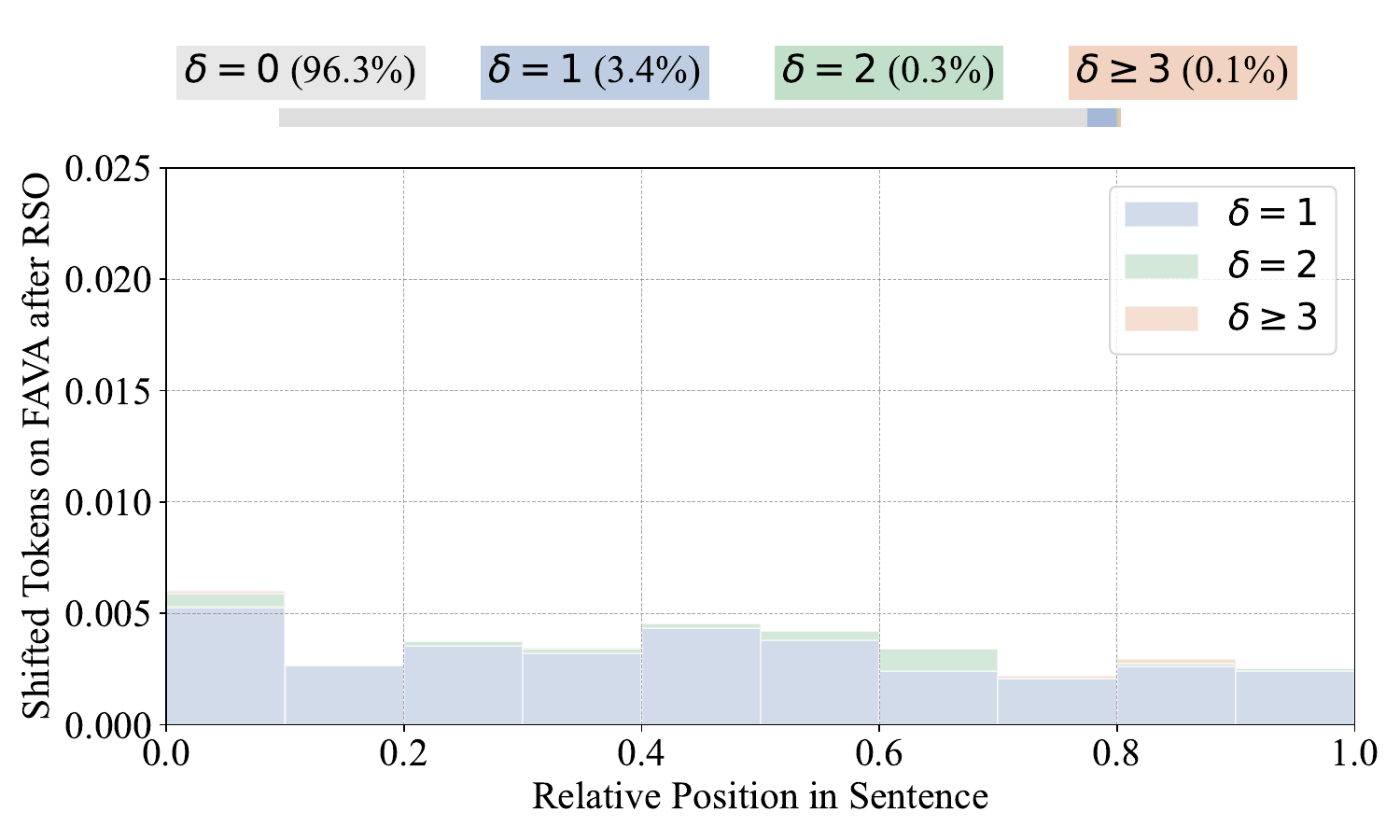}\label{attn_head_influence}} 
\caption{Token shift analysis on LLaMA-2 trained by RSO.}
\label{token_shift_result_rso}
\end{figure}

In this section, we present the results of token analysis of LLaMA-2-7B-Chat trained by IPO (Figure \ref{token_shift_result_ipo}), KTO (Figure \ref{token_shift_result_kto}), CPO (Figure \ref{token_shift_result_cpo}) and RSO (Figure \ref{token_shift_result_rso}).

\section{Additional Results of the Data Quality Experiments}
\label{sec:data_quality_results}

\begin{figure}[]
\begin{center}
\resizebox{\linewidth}{!}{
\includegraphics[]{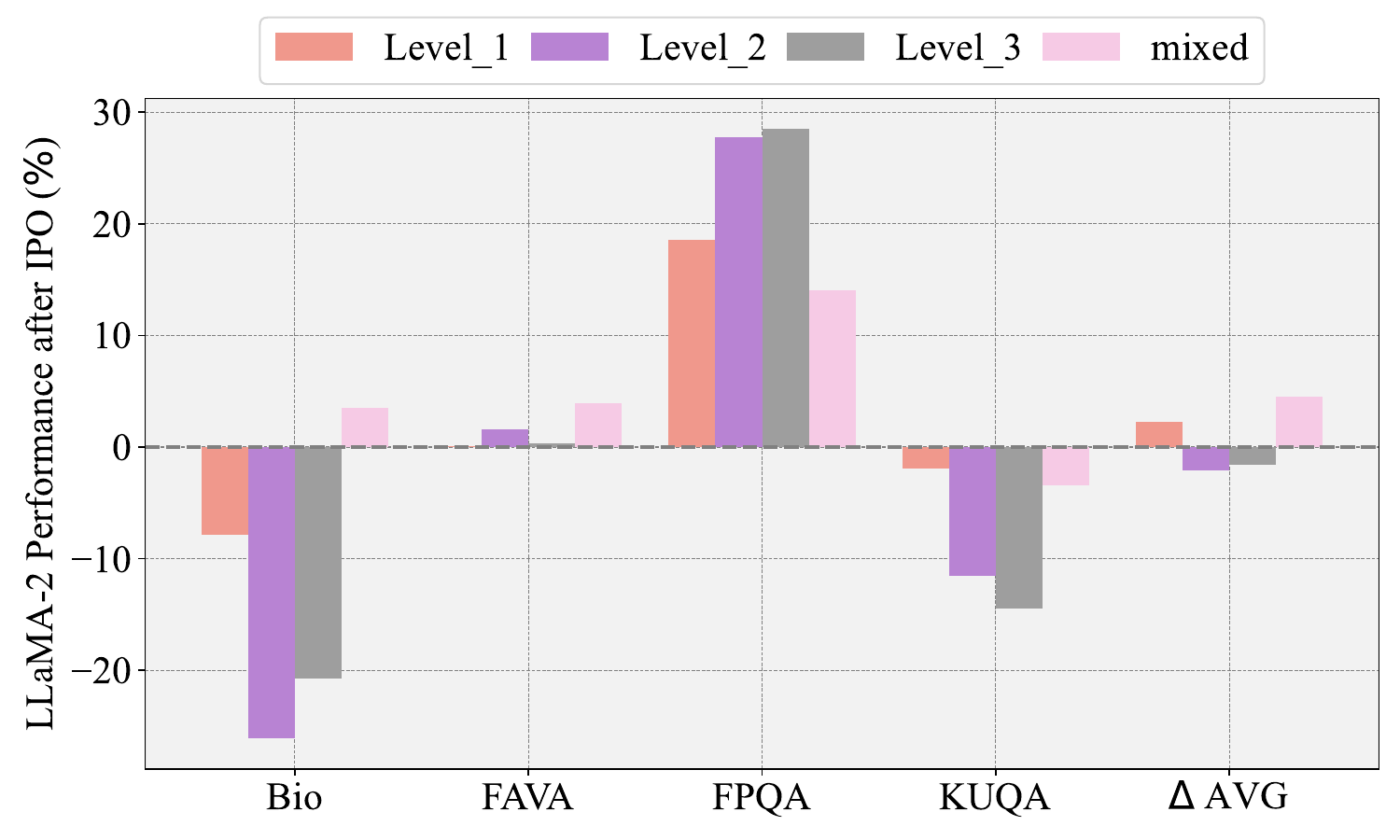} 
}
\caption{Experimental results of models trained by preferences of different quality levels using IPO.}
\label{quality_ipo}
\end{center}
\end{figure}

\begin{figure}[]
\begin{center}
\resizebox{\linewidth}{!}{
\includegraphics[]{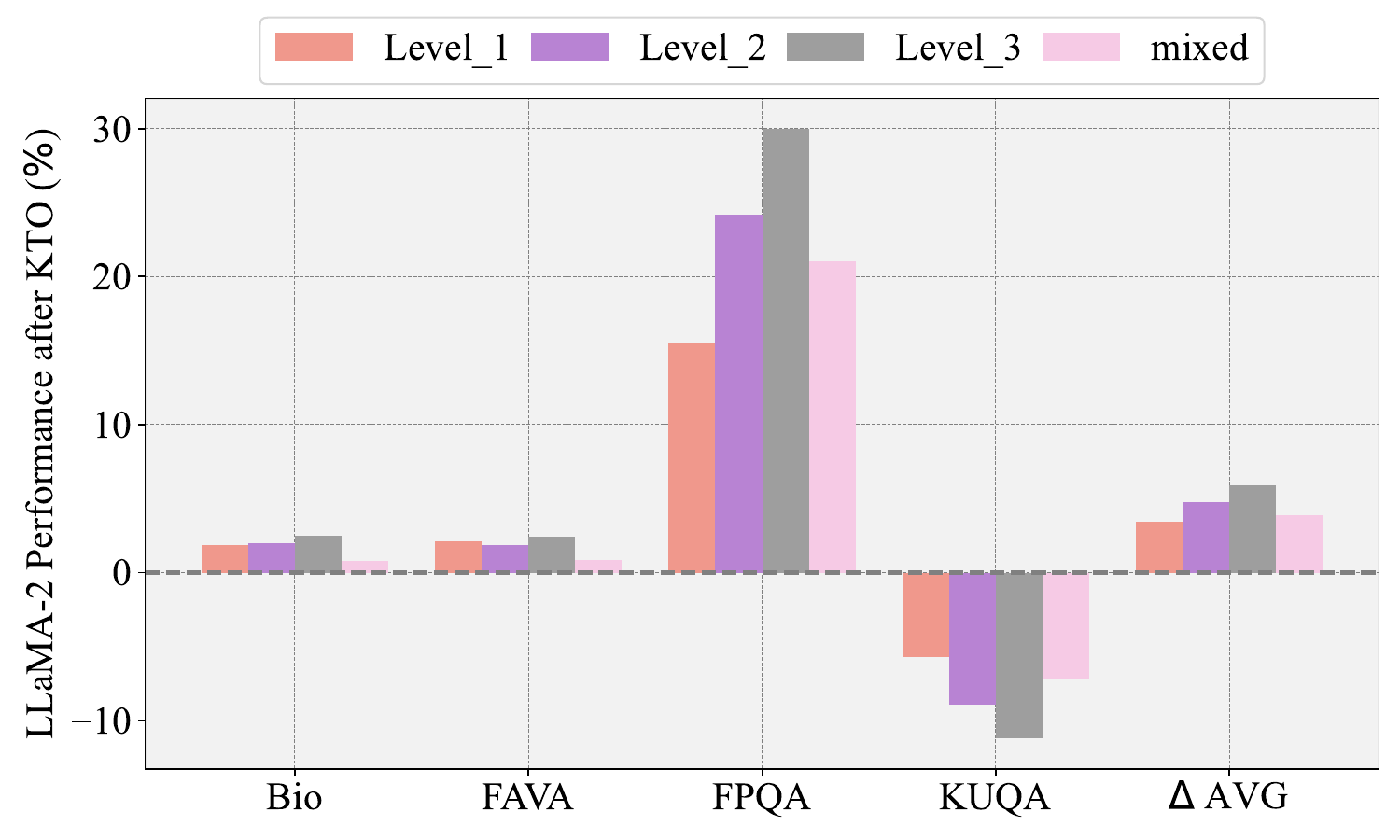} 
}
\caption{Experimental results of models trained by preferences of different quality levels using KTO.}
\label{quality_kto}
\end{center}
\end{figure}

\begin{figure}[]
\begin{center}
\resizebox{\linewidth}{!}{
\includegraphics[]{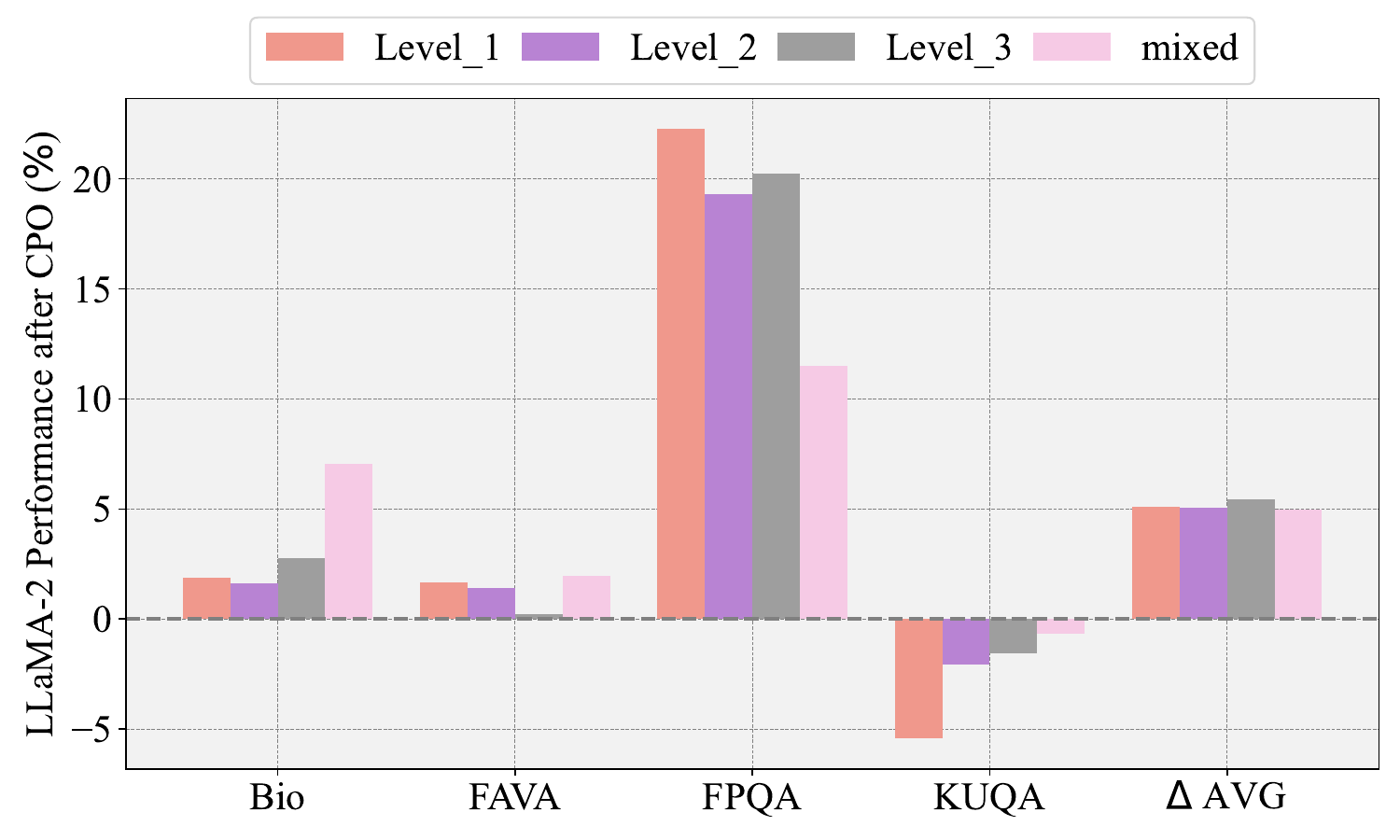} 
}
\caption{Experimental results of models trained by preferences of different quality levels using CPO.}
\label{quality_cpo}
\end{center}
\end{figure}

\begin{figure}[]
\begin{center}
\resizebox{\linewidth}{!}{
\includegraphics[]{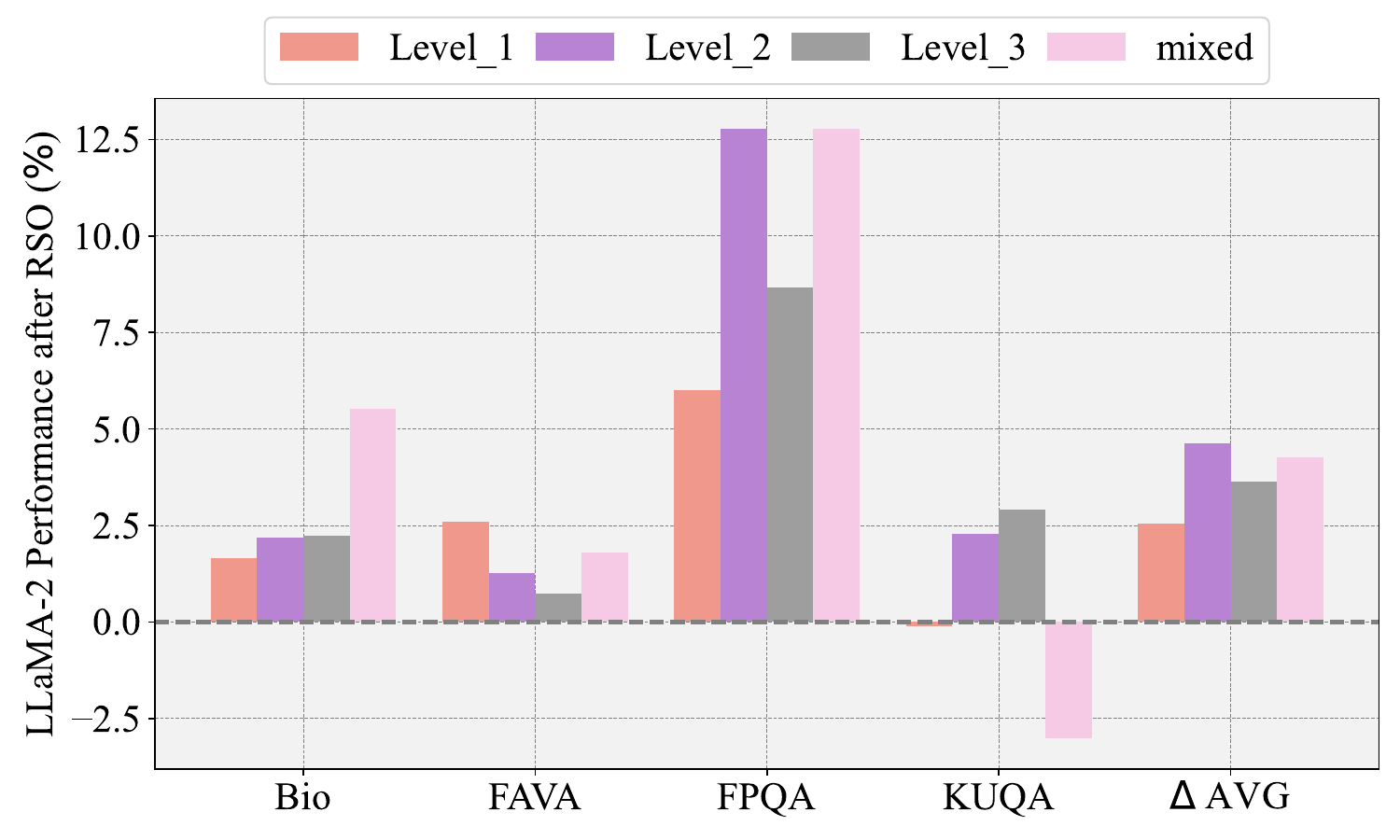} 
}
\caption{Experimental results of models trained by preferences of different quality levels using RSO.}
\label{quality_rso}
\end{center}
\end{figure}

In this section, we present the results demonstrating the impact of data quality on our experiments. The training methods are  IPO (Figure \ref{quality_ipo}), KTO (Figure \ref{quality_kto}), CPO (Figure \ref{quality_cpo}) and RSO (Figure \ref{quality_rso}).

\end{document}